\documentclass[final,times,5p,twocolumn]{elsarticle}

\journal{Computer Vision and Image Understanding}

\usepackage{array}
\usepackage{url}
\usepackage{amssymb}
\usepackage{amsmath}
\usepackage{amsthm}
\usepackage{verbatim}
\usepackage{multirow}
\usepackage{makecell}
\usepackage{tabularx}
\usepackage{graphicx}
\usepackage{subfigure}
\usepackage{subfloat}

\begin{document}
\begin{frontmatter}
\title{Space-Time Representation of People Based on 3D Skeletal Data: A Review}

\author{Fei Han\fnref{equal}}
\ead{fhan@mines.edu}
\author{Brian Reily\fnref{equal}}
\ead{breily@mines.edu}
\author{William Hoff}
\ead{whoff@mines.edu}
\author{Hao Zhang}
\ead{hzhang@mines.edu}

\address{
Division of Computer Science,\\
Colorado School of Mines, Golden, CO 80401, USA
}
\fntext[equal]{These authors contributed equally to this work.}

\begin{abstract}
Spatiotemporal human representation based on 3D visual perception data
is a rapidly growing research area.
Representations can be broadly categorized into two groups, depending on whether they use RGB-D information or 3D skeleton data.
Recently, skeleton-based human representations have been intensively studied and kept attracting an increasing attention,
due to their robustness to variations of viewpoint, human body scale and motion speed as well as the realtime, online performance.
This paper presents a comprehensive survey of existing space-time representations of people based on 3D skeletal data,
and provides an informative categorization and analysis of these methods from the perspectives,
including information modality, representation encoding, structure and transition, and feature engineering.
We also provide a brief overview of skeleton acquisition devices and construction methods, enlist a number of benchmark datasets with skeleton data, and discuss potential future research directions.
\end{abstract}


\end{frontmatter}

\section{Introduction}


Human representation in spatiotemporal space is
a fundamental research problem extensively
investigated in computer vision and machine intelligence
over the past few decades.
The objective of building human representations is to extract compact, descriptive information (i.e., features) to encode and characterize a human's attributes from perception data
(e.g., human shape, pose, and motion),
when developing recognition or other human-centered reasoning systems.
As an integral component of reasoning systems,
approaches to construct human representations have been widely used
in a variety of real-world applications,
including video analysis \cite{ge2012vision},
surveillance \cite{jun2013local},
robotics \cite{demircan2015human},
human-machine interaction \cite{han2017minimum},
augmented and virtual reality \cite{green2007human},
assistive living \cite{okada2005humanoid},
smart homes \cite{brdiczka2009detecting},
education \cite{mondada2009puck},
and many others \cite{Meng_HumanPoseEstimation,broadbent2009acceptance,kang2005hands,fujita2000digital}.


During recent years,
human representations based on 3D perception data have been attracting an increasing amount of attention \cite{vieira2012stop,Kviatkovsky2014,Siddharth2014,li2015feature}.
Comparing with 2D visual data,
additional depth information provides several advantages.
Depth images provide geometric information of pixels that encode the external surface of the scene in 3D space.
Features extracted from depth images and 3D point clouds
are robust to variations of illumination, scale, and rotation \cite{Aggarwal_PRL14,han2013enhanced}.
Thanks to the emergence of affordable structured-light color-depth sensing technology, such as the Microsoft Kinect \cite{Kinect} and Asus Xtion PRO LIVE \cite{Xtion} RGB-D cameras,
it is much easier and cheaper to obtain depth data.
In addition, structured-light cameras enable us to retrieve the 3D human skeletal information in real time \cite{Shotton_CVPR11},
which used to be only possible when using expensive and complex vision systems (e.g., motion capture systems \cite{tobon2010mocap}), thereby significantly popularizing skeleton-based human representations. 
Moreover, the vast increase in computational power allows
researchers to develop advanced computational algorithms (e.g., deep learning \cite{du2015hierarchical}) to process
visual data at an acceptable speed.
The advancements contribute
to the boom of utilizing 3D perception data to construct reasoning systems
in computer vision and machine learning communities.


Since the performance of machine learning and reasoning methods
heavily relies on the design of data representation \cite{bengio2013representation},
human representations are intensively investigated to address human-centered research problems (e.g., human detection, tracking, pose estimation, and action recognition).
Among a large number of human representation approaches \cite{bualan2007detailed,Ganapathi2010,Rahmani2014ECCV,wang2012robust,belagiannis20143d,burenius20133d},
most of the existing 3D based methods can be broadly grouped into two categories: representations based on local features \cite{le2011learning,hzhang_IROS11}
and skeleton-based representations \cite{han2017simultaneous,xu2013efficient,tang2014latent,sun2015cascaded}.
Methods based on local features detect points of interest in space-time dimensions,
describe the patches centered at the points as features,
and encode them (e.g., using bag-of-word models) into representations,
which can locate salient regions and are relatively robust to partial occlusion.
However, methods based on local features ignore spatial relationships among the features.
These approaches are often incapable of identifying feature affiliations, and thus the methods are generally incapable to represent multiple individuals in the same scene.
These methods are also computationally expensive because of the complexity of the procedures including keypoint detection,
feature description, dictionary construction, etc.

\begin{figure}[htb]
\centering
\includegraphics[width= 0.45\textwidth]{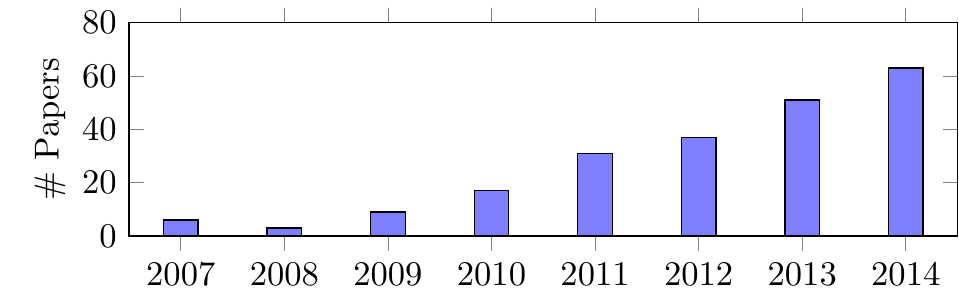}
\caption{Number of 3D skeleton-based human representations published in recent years according to our comprehensive review.}\label{fig:statistics}
\end{figure}

On the other hand,
human representations based on 3D skeleton information
provide a very promising alternative.
The concept of skeleton-based representation can be traced back to the early seminal research of
Johansson \cite{Johansson_PP73},
which demonstrated that a small number of joint positions can effectively represent human behaviors.
3D skeleton-based representations also demonstrate promising performance in real-world applications including Kinect-based gaming, as well as in computer vision research \cite{Yao_BMVC11,du2015hierarchical}.
3D skeleton-based representations are able to model the relationship of human joints and encode the whole body configuration.
They are also robust to scale and illumination changes,
and can be invariant to camera view as well as human body rotation and motion speed.
In addition,
many skeleton-based representations can be computed at a high frame rate,
which can significantly facilitate online, real-time applications.
Given the advantages and previous success of 3D skeleton-based representations,
we have witnessed a significant increase of new techniques to construct such representations in recent years,
as demonstrated in Fig. \ref{fig:statistics},
which underscores the need of this survey paper
focusing on the review of 3D skeleton-based human representations.

%

Several survey papers were published in related research areas
such as motion and activity recognition.
For example,
Han et al. \cite{han2013enhanced} described the Kinect sensor and its general application in computer vision and machine intelligence.
Aggarwal and Xia \cite{Aggarwal_PRL14} recently published a review paper on
human activity recognition from 3D visual data,
which summarized five categories of representations based on
3D silhouettes,
skeletal joints or body part locations,
local spatio-temporal features, scene flow features,
and local occupancy features.
Several earlier surveys were also published to review
methods to recognize human poses, motions, gestures, and activities
\cite{lun2015survey,ruffieux2014survey,borges2013video,Chen_PRL13,ke2013review,LaViola2013,Ye_TOF13,aggarwal2011human,ji2010advances,poppe2010survey,moeslund2006survey,moeslund2001survey},
as well as their applications \cite{chaaraoui2012review,zhou2008human}.
However, none of the survey papers specifically focused on 3D human representation based on skeletal data,
which was the subject of numerous research papers in the literature
and continues to gain popularity in recent years.



The objective of this survey is to provide a comprehensive overview of
3D skeleton-based human representations mainly published in the computer vision and machine intelligence communities,
which are built upon 3D human skeleton data that is assumed as the raw measurements directly from sensing hardware.
We categorize and compare the reviewed approaches from multiple perspectives,
including information modality, representation coding, structure and transition, and
feature engineering methodology, and analyze the pros and cons of each category.
Compared with the existing surveys,
the main contributions of this review include:
\begin{itemize}
\item To the best of our knowledge, this is the first survey dedicated to
\emph{human representations based on 3D skeleton data},
which fills the current void in the literature.

\item The survey is \emph{comprehensive} and covers the \emph{most recent and advanced} approaches.
    We review 171 3D skeleton-based human representations, including 150 papers that were published in the recent five years,
    thereby providing readers with the complete, state-of-the-art methods.
\item This paper provides an insightful categorization and analysis of the 3D skeleton-based representation construction approaches from multiple perspectives,
    and summarizes and compares attributes of all reviewed representations.
\end{itemize}

In addition, we provide a complete list of available benchmark datasets.
Although we also provide a brief overview of human modeling methods to generate skeleton data through pose recognition and joint estimation \cite{zhou2014spatio,akhter2015pose,lehrmann2013non,pons2014posebits},
the purpose is to provide related background information.
Skeleton construction, which is widely studied in the research fields
(such as computer vision, computer graphics, human-computer interaction, and animation) is not the focus of this paper.
In addition, the main application domains of interest in this survey paper is human gesture, action, and activity recognition,
as most of the reviewed papers focus on these applications.
Although several skeleton-based representations are also used for human re-identification \cite{munaro2014one,giachetti2016shrec}, however, skeleton-based features are usually used along with other shape or texture based features (e.g., 3D point cloud) in this application, as skeleton-based features are generally incapable to represent human appearance that is critical for human re-identification.


The remainder of this review is structured as follows.
Background information including 3D skeleton acquisition and construction as well as public benchmark datasets is presented in Section \ref{sec:Background}.
Sections \ref{sec:InfoViews} to \ref{sec:Construction} discuss the categorization of 3D skeleton-based human representations from four perspectives,
including information modality in Section \ref{sec:InfoViews},
encoding in Section \ref{sec:RepEncoding},
hierarchy and transition in Section \ref{sec:Hierarchy},
and feature construction methodology in Section \ref{sec:Construction}.
After discussing the advantages of skeleton-based representations and
pointing out future research directions in Section \ref{sec:discussion},
the review paper is concluded in Section \ref{sec:Conclusion}.

\section{Background} \label{sec:Background}

The objective of building 3D skeleton-based human representations
is to extract compact, discriminative descriptions to characterize a human's attributes from 3D human skeletal information.
The 3D skeleton data encodes human body as an articulated system of rigid segments connected by joints.
This section discusses how 3D skeletal data can be acquired,
including devices that directly provide the skeletal data
and computational methods to construct the skeleton.
Available benchmark datasets including 3D skeleton information are also summarized in this section.

\newcommand{\tabincell}[2]{\begin{tabular}{@{}#1@{}}#2\end{tabular}}

\begin{table*}[htbp]
\centering
\caption{Summary of Recent Skeleton Construction Techniques Based on Depth and/or RGB Images.
}
\label{tab:construction}
\small
\begin{tabular}{|c|c|c|c|}
\hline
Ref. & Approach & Input Data & Performance\\
\hline\hline
\cite{Shotton_CVPR11},\cite{Girshick2011} & Pixel-by-pixel classification & Single depth image & \tabincell{c}{3D skeleton, 16 joints,\\ real-time, 200 fps}\\\hline
\cite{ye2011accurate} & Motion exemplars & Single depth image & 3D skeleton, 38mm accuracy\\\hline
\cite{yub2015random} & Random tree walks & Single depth image & 3D skeleton, real-time, 1000fps\\\hline
\cite{sun2012conditional} & Conditional regression forests & Single depth image & 3D skeleton, over 80\% precision \\\hline
\cite{Charles2011} & Limb-based shape models & Single depth image & 2D skeleton, robust to occlusions \\\hline
\cite{Holt2011} & \tabincell{c}{Decision tree poselets with\\ pictorial structures prior} & Single depth image & \tabincell{c}{3D skeleton, only need small\\ amount of training data} \\\hline
\cite{Grest2005} & ICP using optimized Jacobian & Single depth image & 3D skeleton, over 10 fps\\\hline
\cite{Baak2011} & Matching previous joint positions & Single depth image & \tabincell{c}{3D skeleton, 20 joints, 100 fps,\\ robust to noise and occlusions} \\\hline
\cite{taylor2012vitruvian} & \tabincell{c}{Regression to\\ predict correspondences} & \tabincell{c}{Multiple silhouette \&\\ single depth images} & \tabincell{c}{3D skeleton, 19 joints,\\ real-time, 120fps} \\\hline
\cite{Zhu2008} & ICP on individual parts & Depth image sequence & \tabincell{c}{3D skeleton, 10fps,\\ robust to occlusion} \\\hline
\cite{ganapathi2012real} & ICP with physical constraints & Depth image sequence & \tabincell{c}{3D skeleton, real-time, 125fps,\\ robust to self collision}\\\hline
\cite{Plagemann2010},\cite{Ganapathi2010} & Haar features and Bayesian prior & Depth image sequence & 3D skeleton, real-time \\\hline
\cite{zhang2013unsupervised} & \tabincell{c}{3D non-rigid matching based on\\ MRF deformation model} & Depth image sequence & 3D skeleton\\\hline
\cite{Schwarz2012} & Geodesic distance \& optical flow & RGBD image streams &  \tabincell{c}{3D skeleton, 16 joints,\\ robust to occlusions}\\\hline
\cite{ichim2016semantic} & \tabincell{c}{energy cost optimization\\with multi-constraints} & Depth image sequence & model per frame = 3ms\\\hline
\hline
\cite{ionescu2014iterated} & Second-order label-sensitive pooling & RGB images & 3D pose, 106mm precision\\\hline
\cite{Anon2014} & Recurrent 2D/3D pose estimation & Single RGB images & \tabincell{c}{3D skeleton, robust to viewpoint\\ changes and occlusions} \\\hline
\cite{fan2015combining} & Dual-source deep CNN & Single RGB images & 2D skeleton, robust to occlusions\\\hline
\cite{toshev2014deeppose} & Deep neural networks & Single RGB images & \tabincell{c}{2D skeleton, robust to\\ appearance variations}\\\hline
\cite{dong2014towards} & Parselets/grid layout feature & Single RGB images & 2D skeleton, robust to occlusions \\\hline
\cite{akhter2015pose} & Prior based on joint angle limits & Single RGB images & 3D skeleton \\\hline
\cite{tompson2014joint} & CNN/Markov random field & Single RGB images & 2D skeleton, close to real-time \\\hline
\cite{elhayek2015efficient} & ConvNet joint detector & \tabincell{c}{Multi-perspective\\ RGB images} & \tabincell{c}{2D skeleton,\\  nearly 95\% accuracy}\\\hline
\cite{gall2009motion},\cite{liu2011markerless} &
\tabincell{c}{Skeleton tracking and \\ surface estimation} & \tabincell{c}{Multi-perspective\\ RGB images}  & \tabincell{c}{3D skeleton, deal with rapid\\ movements \& apparel like skirts} \\\hline


\end{tabular}
\end{table*}


\subsection{Direct Acquisition of 3D Skeletal Data}

\begin{figure}[htbp]
\centering
\includegraphics[width=0.49\textwidth]{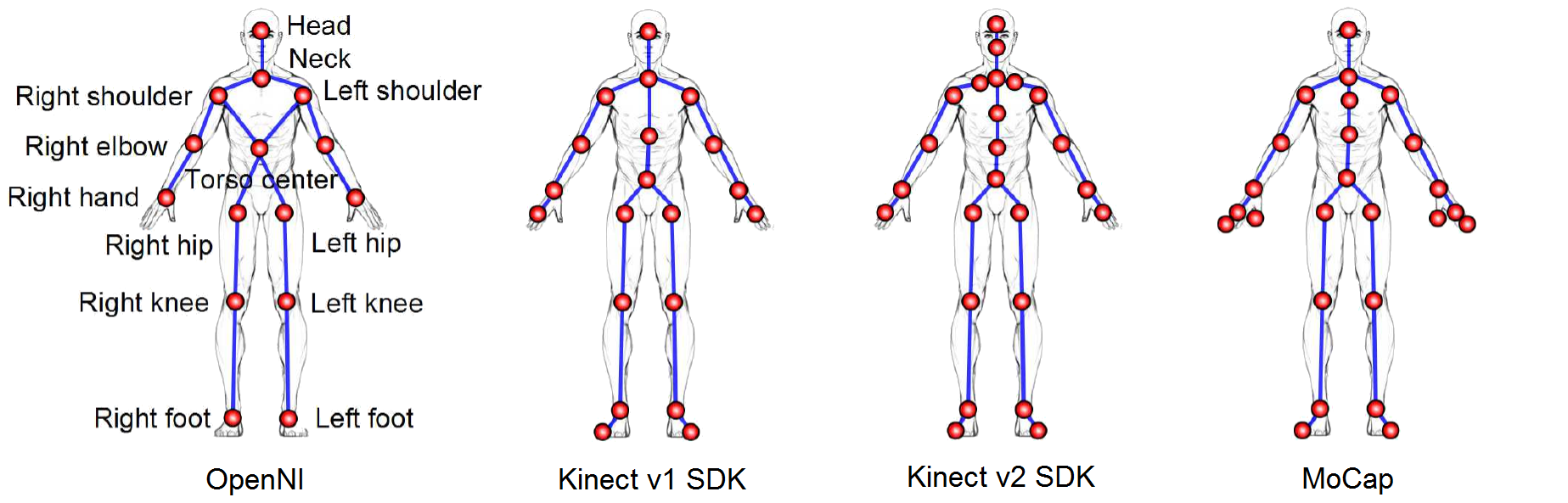}
\caption{
Examples of skeletal human body models obtained
from different devices.
The OpenNI library tracks 15 joints;
Kinect v1 SDK tracks 20 joints;
Kinect v2 SDK tracks 25;
and MoCap systems can track various numbers of joints.
}\label{fig:skeleton}
\end{figure}

Several commercial devices, including motion capture systems, time-of-flight sensors, and structured-light cameras,
allow for direct retrieval of 3D skeleton data.
The 3D skeletal kinematic human body models provided by the devices are shown in Fig. \ref{fig:skeleton}.

\subsubsection{Motion Capture Systems (MoCap)}

Motion capture systems identify and track markers
that are attached to a human subject's joints or body parts to obtain 3D skeleton information.
There are two main categories of MoCap systems, based on either visual cameras or inertia sensors.
Optical-based systems employ multiple cameras positioned around a subject
to track, in 3D space, reflective markers attached to the human body.
In MoCap systems based on inertial sensors,
each 3-axis inertial sensor estimates the rotation of
a body part with respect to a fixed point.
This information is collected to obtain the skeleton data without any optical devices around
a subject.
Software to collect skeleton data is provided with commercial MoCap systems,
such as Nexus for Vicon MoCap\footnote{Vicon: \url{http://www.vicon.com/products/software/nexus}.},
NatNet SDK for OptiTrack\footnote{OptiTrack: \url{http://www.optitrack.com/products/natnet-sdk}.}, etc.
MoCap systems, especially based on multiple cameras,
can provide very accurate 3D skeleton information at a very high speed.
On the other hand, such systems are typically expensive and
can only be used in well controlled indoor environments.


\subsubsection{Structured-Light Cameras}

Structured-light color-depth sensors are a type of camera
that uses infrared light to capture depth information about a scene, such as
Microsoft Kinect v1 \cite{Kinect},
ASUS Xtion PRO LIVE \cite{Xtion},
and PrimeSense \cite{primesense}, among others.
A structured-light sensor consists of an infrared-light source and a receiver
that can detect infrared light.
The light projector emits a known pattern,
and the way that this pattern distorts on the scene allows the camera to decide the depth.
A color camera is also available on the sensor to acquire color frames that can be registered to depth frames,
thereby providing color-depth information at each pixel of a frame or 3D color point clouds.
Several drivers are available to provide the access to the color-depth data acquired by the sensor,
including the Microsoft Kinect SDK \cite{Kinect}, the OpenNI library
\cite{openni},
and the OpenKinect library \cite{openkinect}.
The Kinect SDK also provides 3D human skeletal data
using the method described by Shotton et.al \cite{Shotton2011}.
OpenNI uses NITE \cite{nite} -- a skeleton generation framework developed as proprietary software by PrimeSense, to generate a similar 3D human skeleton model.
Markers are not necessary for structured-light sensors.
They are also inexpensive and can provide 3D skeleton information in real time.
On the other hand, since structured-light cameras are based on infrared light,
they can only work in an indoor environment.
The frame rate (30 Hz) and resolution of depth images ($320\!\times\! 240$) are also relatively low.

\subsubsection{Time-of-Flight (ToF) Sensors}

ToF sensors are able to acquire accurate depth data at a high frame rate,
by emitting light and measuring the amount of time it takes for that light to return
-- similar in principle to
established depth sensing technologies, such as radar and LiDAR.
Compared to other ToF sensors,
the Microsoft Kinect v2 camera offers an affordable alternative to acquire depth data using this technology.
In addition, a color camera is integrated into the sensor to provide registered color data.
The color-depth data can be accessed by the Kinect SDK 2.0 \cite{kinect2}
or the OpenKinect library (using the libfreenect2 driver) \cite{openkinect}.
The Kinect v2 camera provides a higher resolution of depth images ($512 \! \times \! 424$) at 30 Hz.
Moreover,
the camera is able to provide 3D skeleton data by estimating positions of 25 human joints,
with better tracking accuracy than the Kinect v1 sensor.
Similar to the first version, the
Kinect v2 has a working range of approximately 0.5 to 5 meters.


\subsection{3D Pose Estimation and Skeleton Construction}

Besides manual human skeletal joint annotation \cite{lee2005dynamic,Holt2011,Johnson10},
a number of approaches have been designed to automatically construct a skeleton model from perception data through pose recognition and joint estimation.
Some of these are based on methods used in RGB imagery,
while others take advantage of the extra information available in a depth or RGB-D image.
The majority of the current methods are based on body part recognition, and
then fit a flexible model to the now `known' body part locations.
An alternate main methodology is starting with a `known' prior, and fitting
the silhouette or point cloud to this prior after the humans are localized \cite{Ikemura2010,Spinello2011,hzhang_IROS11}.
This section provides a brief review of autonomous skeleton construction methods based on visual data according to the information that is used.
A summary of the reviewed skeleton construction techniques is presented in Table \ref{tab:construction}.

\subsubsection{Construction from Depth Imagery} \label{sec:sub:ConstructDepth}

Due to the additional 3D geometric information that depth imagery can provide,
many methods are developed to build a 3D human skeleton model based on
a single depth image or a sequence of depth frames.

Human joint estimation via body part recognition is one popular
approach to construct the skeleton model \cite{Shotton_CVPR11,
Girshick2011, Plagemann2010, Holt2011, Charles2011, Schwarz2012,
sun2012conditional, yub2015random}.
A seminal paper by Shotton et al. \cite{Shotton_CVPR11} in 2011 provided an
extremely effective skeleton construction algorithm based on body part
recognition, that was able to work in real time.
A single depth image (independent of previous frames) is classified on a
per-pixel basis, using a randomized decision forest classifier.
Each branch in the forest is determined by a simple relation between the target
pixel and various others.
The pixels that are classified into the same category form the body part,
and the joint is inferred by the mean-shift method from a certain body part,
using the depth data to `push' them into the silhouette.
While training the decision forests takes a large number of images
(around 1 million) as well as a considerable amount of computing power,
the fact that the branches in the forest are very simple allows this algorithm
to generate 3D human skeleton models within about 5 ms.
An extended work was published in \cite{Girshick2011},
with both accuracy and speed improved.
Plagemann et al. \cite{Plagemann2010} introduced an approach to recognize body parts
using Haar features \cite{Gould2010} and construct a skeleton model on these parts.
Using data over time, they construct a Bayesian network, which produces
the estimated pose using body part locations and starts with the
previous pose as a prior \cite{Ganapathi2010}.
Holt et al. \cite{Holt2011} proposed Connected Poselets to estimate 3D human pose from depth data.
The approach utilizes the idea of poselets \cite{Bourdev2009},
which is widely applied for pose estimation from RGB images.
For each depth image, a multi-scale sliding window is applied,
and a decision forest is applied to detect poselets and estimate human joint locations.
Using a skeleton prior inspired by pictorial structures \cite{Fischler1973,Andriluka2009},
the method begins with a torso point and connects outwards to body parts.
By applying kinematic inference to eliminate impossible poses,
they are able to reject incorrect body part classifications and
improve their accuracy.

Another widely investigated methodology to construct 3D human skeleton models from depth imagery
is based on nearest-neighbor matching \cite{Grest2005,Zhu2008,Baak2011,taylor2012vitruvian,zhang2013unsupervised,ye2011accurate}. 
Several approaches for whole-skeleton matching are based on the Iterative
Closest Point (ICP) method \cite{Besl1992},
which can iteratively decide a rigid transformation
such that the input query points fit to the points in the given model under this transformation.
Using point clouds of a person with known poses as a model,
several approaches \cite{Grest2005,Zhu2008} apply ICP to fit the unknown poses by
estimating the translation and rotation to fit the unknown body parts to the known model.
While these approaches are relatively accurate,
they suffer from several drawbacks.
ICP is computationally expensive for a model with as many degrees of freedom as a human body.
Additionally, it can be difficult to recover from tracking loss.
Typically the previous pose is used as the known pose to fit to; if
tracking loss occurs and this pose becomes inaccurate, then further
fitting can be difficult or impossible.
Finally, skeleton construction methods based on the ICP algorithm generally require an initial T-pose to start the iterative process.

\subsubsection{Construction from RGB Imagery}

Early approaches and several recent methods based on deep learning
focused on 2D or 3D human skeleton construction from traditional RGB or intensity images,
typically by identifying human body parts using visual features (e.g., image gradients, deeply learned features, etc.),
or matching known poses to a segmented silhouette.

\textbf{\emph{Methods based on a single image}}:
Many algorithms were proposed to construct human skeletal model
using a single color or intensity image acquired from a monocular camera \cite{Anon2014,Yang2011,dong2014towards,akhter2015pose}.
Wang et al. \cite{Anon2014} constructs a 3D human skeleton from a single image using a linear combination
of known skeletons with physical constraints on limb lengths.
Using a 2D pose estimator \cite{Yang2011},
the algorithm begins with a known 2D pose and a mean 3D pose,
and calculates camera parameters from this estimation.
The 3D joint positions are recalculated using the estimated parameters, and
the camera parameters are updated.
The steps continue iteratively until convergence.
This approach was demonstrated to be robust to partial occlusions and errors in the 2D estimation.
Dong et al. \cite{dong2014towards} considered the human parsing and pose estimation
problems simultaneously.
The authors introduced a unified framework based on
semantic parts using a tailored And-Or graph.
The authors also employed parselets and Mixture of Joint-Group Templates
as the representation.
%

Recently, deep neural networks have proven their ability in human skeleton
construction \cite{toshev2014deeppose,tompson2014joint,fan2015combining}.
Toshev and Szegedy \cite{toshev2014deeppose} employed Deep Neural Networks (DNNs)
for human pose estimation.
The proposed cascade of DNN regressors obtains pose estimation results
with high precision.
Fan et al. \cite{fan2015combining} uses Dual-Source Deep Convolutional
Neural Networks (DS-CNNs) for estimating 2D human poses from a single image.
This method takes a set of image patches as the input
and learns the appearance of each local body part
by considering their previous views in the full body,
which successfully addresses the joint recognition and localization issue.
Tompson et al. \cite{tompson2014joint} proposed a unified learning
framework based on deep Convolutional Networks (ConvNets) and
Markov Random Fields,
which can generate a heat-map to encode a per-pixel
likelihood for human joint localization from a single RGB image.

\textbf{\emph{Methods based on multiple images:}}
When multiple images of a human are acquired from different perspectives by
a multi-camera system,
traditional stereo vision techniques can be employed to estimate depth maps of the human.
After obtaining the depth image,
a human skeleton model can be constructed using methods based on depth information (Section \ref{sec:sub:ConstructDepth}).
Although there exists a commercial solution that uses
marker-less multi-camera systems to obtain highly
precise skeleton data at 120
frames per second (FPS) and approximately 25-50ms latency \cite{brooks2012markerless},
computing depth maps is usually slow and often suffers from problems
such as failures of correspondence search and noisy depth information.
To address these problems,
algorithms were also studied to construct human skeleton models directly from
the multi-images without calculating the depth image \cite{elhayek2015efficient,gall2009motion,liu2011markerless}.
For example,
Gall et al. \cite{gall2009motion} introduced an approach to fully-automatically estimate
the 3D skeleton model from a multi-perspective video sequence, where an articulated
template model and silhouettes are obtained from the sequence.
Another method was also proposed by Liu et al. \cite{liu2011markerless},
which uses a modified global optimization method to handle occlusions.

\subsection{Benchmark Datasets With Skeletal Data}

In the past five years, a large number of benchmark datasets containing 3D human skeleton data were collected in different scenarios and made available to the public.
This section provides a complete review of the datasets as listed in Table \ref{tab:dataset}.
We categorize and discuss these datasets according to the type of devices used to acquire the skeleton information.

\subsubsection{Datasets Collected Using MoCap Systems}

Early 3D human skeleton datasets were usually collected by a MoCap system,
which can provide accurate locations of a various number of skeleton joints
by tracking the markers attached on human body,
typically in indoor environments.
The CMU MoCap dataset \cite{CMU_MOCAP} is one of the earliest resources that consists of a wide variety of human actions,
including interaction between two subjects, human locomotion,
interaction with uneven terrain, sports, and other human actions.
It is capable of recording 120 Hz with images of 4 megapixel resolution.
The recent Human3.6M dataset \cite{ionescu2014human3} is one of the largest MoCap datasets, which consists of 3.6 million human poses and corresponding images captured by a high-speed MoCap system.
There are 4 basler high-resolution progressive scan cameras to acquire video data at 50 Hz.
It contains activities by 11 professional actors in 17 scenarios:
discussion, smoking, taking photo, talking on the phone, etc.,
as well as provides accurate 3D joint positions and high-resolution videos.
The PosePrior dataset \cite{akhter2015pose} is the newest MoCap dataset
that includes an extensive variety of
human stretching poses performed by trained athletes and gymnasts.
Many other MoCap datasets were also released,
including the
Pictorial Human Spaces  \cite{marinoiu2013pictorial},
CMU Multi-Modal Activity (CMU-MMAC) \cite{CMU_MMAC}
Berkeley MHAD \cite{ofli2013berkeley},
Standford ToFMCD \cite{Ganapathi2010},
HumanEva-I \cite{sigal2010humaneva}, and
HDM05 MoCap \cite{HDM05} datasets.

\begin{table*}[htbp]
\centering
\caption{Publicly Available Benchmark Datasets Providing 3D Human Skeleton Information.}
\label{tab:dataset}
\small
\begin{tabular}{|c|c|c|c|c|c|c|}
\hline
Year & Dataset and Reference &   Acquisition device  & Other Data & Scenario\\
\hline\hline
2016 & Shrec'16 \cite{giachetti2016shrec} & Xtion Live Pro & Point cloud & person re-identification\\\hline
2015 & $M^2I$ \cite{xu2015multi} & Kinect v1 & RGB + depth & human daily activities\\\hline
2015 & Multi-View TJU \cite{liu2015single} & Kinect v1 & RGB + depth & human daily activities\\\hline
2015 & PosePrior \cite{akhter2015pose} & MoCap & color & extreme motions\\\hline
2015 & SYSU 3D HOI \cite{hu2015jointly} & Kinect v1 & color + depth & human daily activities \\\hline
2015 & TST Intake Monitoring \cite{cippitelli2015comparison} & Kinect v2 + IMU & depth & human daily activities\\\hline
2015 & TST TUG \cite{cippitelli2015time} & Kinect v2 + IMU & depth & human daily activities\\\hline
2015 & UTD-MHAD \cite{chen2015utd} & Kinect v1 + IMU & RGB + depth & atomic actions\\\hline

2014 & BIWI RGBD-ID \cite{munaro2014one} & Kinect v1 & RGB + depth & person re-identification\\\hline
2014 & CMU-MAD \cite{huang2014sequential} & Kinect v1 & RGB + depth & atomic actions\\\hline
2014 & G3Di \cite{bloom2014g3di} & Kinect v1 & RGB + depth & gaming\\\hline
2014 & Human3.6M \cite{ionescu2014human3} & MoCap & color & movies\\\hline
2014 & \tabincell{c}{Northwestern-UCLA\\ Multiview \cite{wang2014cross}} & Kinect v1 & RGB + depth
				& human daily activities\\\hline
2014 & ORGBD \cite{yu2015discriminative} & Kinect v1 & RGB + depth & human-object interactions \\\hline
2014 & SPHERE \cite{paiementonline} & Kinect
				& depth & human daily activities\\\hline
2014 & TST Fall Detection \cite{gasparrini2014}	& Kinect v2 + IMU & depth & human daily activities\\\hline

2013 & Berkeley MHAD \cite{ofli2013berkeley} & MoCap & RGB + depth & human daily activities\\\hline
2013 & CAD-120 \cite{koppula2013learning} & Kinect v1 & RGB + depth & human daily activities\\\hline
2013 & ChaLearn \cite{escalera2013multi} & Kinect v1 & RGB + depth & Italian gestures\\\hline
2013 & KTH Multiview Football \cite{kazemi2013multi} & 3 cameras & color & football activities\\\hline
2013 & MSR Action Pairs \cite{oreifej2013hon4d} & Kinect v1 & RGB + depth & activities in pairs\\\hline
2013 & Multiview 3D Event \cite{Wei2013} & Kinect v1 & RGB + depth & indoor human activities\\\hline
2013 & Pictorial Human Spaces \cite{marinoiu2013pictorial} & MoCap & color & human daily activities\\\hline
2013 & UCF-Kinect \cite{Ellis_IJCV13} & Kinect v1 & color & human daily activities\\\hline

2012 & 3DIG \cite{sadeghipour2012gesture} & Kinect v1 & color + depth & iconic gestures\\\hline
2012 & CAD-60 \cite{Sung_ICRA12} & Kinect v1 & RGB + depth & human daily activities\\\hline
2012 & Florence 3D-Action \cite{Seidenari2013} & Kinect v1 & color
				& human daily activities\\\hline
2012 & G3D \cite{Bloom_CVPRW12}& Kinect v1 & RGB + depth & gaming\\\hline
2012 & MSRC-12 Gesture \cite{MSRC12b} & Kinect v1 & N/A & gaming\\\hline
2012 & MSR Daily Activity 3D \cite{Wang_CVPR12} & Kinect v1 & RGB + depth
              & human daily activities\\\hline
2012 & \tabincell{c}{RGB-D Person\\ Re-identification \cite{Barbosa_reid12}}& Kinect v1 & RGB + 3D mesh & person re-identification\\\hline
2012 & SBU-Kinect-Interaction \cite{Yun_CVPRW12} & Kinect v1 & RGB + depth & human interaction
				activities\\\hline
2012 & UT Kinect Action \cite{Xia_CVPRW12} & Kinect v1 & RGB + depth & atomic actions\\\hline

2011 & CDC4CV pose \cite{Holt2011} & Kinect v1 & depth & basic activities\\\hline

2010 & HumanEva \cite{sigal2010humaneva} & MoCap & color & human daily activities\\\hline
2010 & MSR Action 3D \cite{Wang_CVPR12} & Kinect v1 & depth & gaming\\\hline
2010 & Stanford ToFMCD \cite{Ganapathi2010} & MoCap + ToF & depth & human daily activities\\\hline
2009 & TUM kitchen \cite{tenorth2009tum} & 4 cameras & color & manipulation activities\\\hline
2008 & CMU-MMAC \cite{CMU_MMAC} & MoCap & color & cooking in kitchen\\\hline
2007 & HDM05 MoCap \cite{HDM05} & MoCap & color & human daily activities\\\hline
2001 & CMU MoCap \cite{CMU_MOCAP} & MoCap & N/A & gaming + sports + movies\\
\hline
\end{tabular}
\end{table*}

\subsubsection{Datasets Collected by Structured-Light Cameras}

Affordable structured-light cameras are widely used for 3D human skeleton data acquisition.
Numerous datasets were collected using the Kinect v1 camera in different scenarios.
The MSR Action3D dataset \cite{Li2010,Wang_CVPR12} was captured using the Kinect camera at Microsoft Research,
which consists of subjects performing American Sign Language gestures
and a variety of typical human actions, such as making a phone call or reading
a book.
The dataset provides RGB, depth, and skeleton information generated by the Kinect v1 camera for each data instance.
A large number of approaches used this dataset for evaluation and validation \cite{Lopez2014}.
The MSRC-12 Kinect gesture dataset \cite{MSRC12b,MSRC12a}
is one of the largest gesture databases available.
Consisting of nearly seven hours of data and over 700,000 frames of
a variety of subjects performing different gestures, it provides the pose
estimation and other data that was recorded with a Kinect v1 camera.
The Cornell Activity Dataset (CAD) includes
CAD-60 \cite{Sung_ICRA12} and CAD-120 \cite{koppula2013learning},
which contains 60 and 120 RGB-D videos of human daily activities,
respectively.
The dataset was recorded by a Kinect v1 in different environments,
such as an office, bedroom, kitchen, etc.
The SBU-Kinect-Interaction dataset \cite{Yun_CVPRW12}
contains skeleton data of a pair of subjects performing different interaction activities - one person acting and the other reacting.
Many other datasets captured using a Kinect v1 camera were also released to the public,
including the
MSR Daily Activity 3D \cite{Wang_CVPR12},
MSR Action Pairs  \cite{oreifej2013hon4d},
Online RGBD Action (ORGBD) \cite{yu2015discriminative},
UTKinect-Action \cite{Xia_CVPRW12},
Florence 3D-Action \cite{Seidenari2013},
CMU-MAD \cite{huang2014sequential},
UTD-MHAD \cite{chen2015utd},
G3D/G3Di  \cite{Bloom_CVPRW12,bloom2014g3di},
SPHERE \cite{paiementonline},
ChaLearn \cite{escalera2013multi},
RGB-D Person Re-identification \cite{Barbosa_reid12},
Northwestern-UCLA Multiview Action 3D  \cite{wang2014cross},
Multiview 3D Event  \cite{Wei2013},
CDC4CV pose  \cite{Holt2011},
SBU-Kinect-Interaction  \cite{Yun_CVPRW12},
UCF-Kinect  \cite{Ellis_IJCV13},
SYSU 3D Human-Object Interaction  \cite{hu2015jointly},
Multi-View TJU \cite{liu2015single},
$M^2I$ \cite{xu2015multi},
and
3D Iconic Gesture \cite{sadeghipour2012gesture} datasets.
The complete list of human-skeleton datasets collected using structured-light cameras are presented in Table \ref{tab:dataset}.

\subsubsection{Datasets Collected by Other Techniques}

Besides the datasets collected by MoCap or structured-light cameras,
additional technologies were also applied to collect datasets containing 3D human skeleton information,
such as multiple camera systems, ToF cameras such as the Kinect v2 camera,
or even manual annotation.

Due to the low price and improved performance of the Kinect v2 camera,
it has become increasingly widely adopted to collect 3D skeleton data.
The Telecommunication Systems Team (TST) created
a collection of datasets using Kinect v2 ToF cameras,
 which include three datasets for different purposes.
The TST fall detection dataset \cite{gasparrini2014} contains eleven different subjects performing falling activities and activities of daily living
in a variety of scenarios;
The TST TUG dataset \cite{cippitelli2015time}
contains twenty different individuals standing up and walking around;
and the TST intake monitoring dataset contains  food intake actions  performed by 35 subjects \cite{cippitelli2015comparison}.

Manual annotation approaches are also widely used to provide skeleton data.
The 
KTH Multiview Football dataset \cite{kazemi2013multi}
contains images of professional football players
during real matches, which are obtained using color sensors from 
3 views. There are 14 annotated joints for each frame.
Several other skeleton datasets are collected based on manual annotation, including
the LSP dataset \cite{Johnson10},
and the TUM Kitchen dataset \cite{tenorth2009tum}, etc.

\begin{table*}[htbp]
\centering
\caption{Summary of 3D Skeleton-Based Representations Based on Joint Displacement Features.}
\label{tab:Displacement}
\small
Notation:
In the \emph{feature encoding} column: \underline{Conc}atenation-based encoding,
\underline{Stat}istics-based encoding, \underline{B}ag-\underline{o}f-\underline{W}ords encoding.
In the \emph{structure and transition} column: \underline{Low}-\underline{l}e\underline{v}el features,
\underline{Body} parts models, \underline{Manif}olds;
In the \emph{feature engineering} column: \underline{Hand}-crafted features,
\underline{Dict}ionary learning, \underline{Unsup}ervised feature learning,
\underline{Deep} learning.
In the remaining columns:
`T' indicates that temporal information is used in feature extraction;
`VI' stands for \underline{V}iew-\underline{I}nvariant;
`ScI' stands for \underline{Sc}ale-\underline{I}nvariant;
`SpI' stands for \underline{Sp}eed-\underline{I}nvariant;
`OL' stands for \underline{O}n\underline{L}ine;
`RT' stands for \underline{R}eal-\underline{T}ime.
\vspace{6pt}

\tabcolsep=0.15cm
\begin{tabular}{|c|c|c|c|c||c|c|c|c|c|c|}
\hline
Reference & Approach & \tabincell{c}{Feature \\ Encoding} & \tabincell{c}{Structure \& \\ Transition} & \tabincell{c}{Feature\\ Engineering} & T & VI & ScI & SpI & OL & RT \\
\hline
\hline
Hu et al. \cite{hu2015jointly} & JOULE & BoW & Lowlv & Unsup & \checkmark & \checkmark & \checkmark &  &  &  \\\hline
Wang et al. \cite{wang2014cross} & Cross View & BoW & Body & Dict & \checkmark & \checkmark &  &  &  &  \\\hline
Wei et al. \cite{Wei2013} & 4D Interaction & Conc & Lowlv & Hand & \checkmark & \checkmark &  & \checkmark &  &  \\\hline
Ellis et al. \cite{Ellis_IJCV13} & Latency Trade-off & Conc & Lowlv & Hand & \checkmark & \checkmark &  &  & \checkmark & \checkmark \\\hline
Wang et al. \cite{Wang_CVPR12, Wang_PAMI14} & Actionlet & Conc & Lowlv & Hand &  & \checkmark & \checkmark & \checkmark &  &  \\\hline
Barbosa et al. \cite{Barbosa_reid12} & Soft-biometrics Feature & Conc & Body & Hand &  &  &  &  &  &  \\\hline
Yun et al. \cite{Yun_CVPRW12} & Joint-to-Plane Distance & Conc & Lowlv & Hand & \checkmark & \checkmark &  &  &  & \checkmark \\\hline
Yang and Tian \cite{Yang_CVPRW12}, \cite{Yang_JVCIR13} & EigenJoints & Conc & Lowlv & Unsup & \checkmark & \checkmark & \checkmark &  & \checkmark & \checkmark \\\hline
Chen and Koskela \cite{chen2013online} & Pairwise Joints & Conc & Lowlv & Hand &  & \checkmark &  &  & \checkmark & \checkmark \\\hline
Rahmani et al. \cite{Rahmani_WACV14} & Joint Movement Volumes & Stat & Lowlv & Hand &  &  &  &  &  & \checkmark \\\hline
Luo et al. \cite{Luo_ICCV13} & Sparse Coding & BoW & Lowlv & Dict &  & \checkmark &  & \checkmark &  &  \\\hline
Jiang et al. \cite{Jiang2014} &  Hierarchical Skeleton  & BoW & Lowlv & Hand & \checkmark & \checkmark &  &  & \checkmark & \checkmark \\\hline
Yao and Li \cite{yao2012action} & 2.5D Graph Representation  & BoW & Lowlv & Hand & & \checkmark & \checkmark & & &\\\hline
Vantigodi and Babu \cite{vantigodi2013real} & Variance of Joints  & Stat & Lowlv & Hand & \checkmark & \checkmark &  &  &  &  \\\hline
Zhao et al. \cite{zhao2013online} & Motion Templates  & BoW & Lowlv & Dict &  & \checkmark & \checkmark &  & \checkmark & \checkmark \\\hline
Yao et al. \cite{Yao2012} & Coupled Recognition  & Conc & Lowlv & Hand & \checkmark &  &  &  &  &  \\\hline
Zhang et al. \cite{Zhang12} & Star Skeleton & BoW & Lowlv & Hand & \checkmark & \checkmark &  & \checkmark & \checkmark & \checkmark \\\hline
Zou et al. \cite{Zou13} & Key Segment Mining  & BoW & Lowlv & Dict & \checkmark & \checkmark & \checkmark &  &  &  \\\hline
\tabincell{c}{Kakadiaris and\\ Metaxas \cite{Kakadiaris2000}} & Physics Based Model & Conc & Lowlv & Hand & \checkmark &  &  &  &  &  \\\hline
Nie et al. \cite{xiaohan2015joint} & ST Parts & BoW & Body & Dict & \checkmark & \checkmark &  &  &  &  \\\hline
Anirudh et al. \cite{anirudh2015elastic} & TVSRF Space  & Conc & Manif & Hand & \checkmark & \checkmark & \checkmark & \checkmark &  &  \\\hline
Koppula and Saxena \cite{koppula2013learningICML} & Temporal Relational Features  & Conc & Lowlv & Hand & \checkmark &  &  &  &  &  \\\hline
Wu and Shao \cite{Wu2014} & EigenJoints  & Conc & Lowlv & Deep & \checkmark & \checkmark & \checkmark &  &  & \checkmark \\\hline
Kerola et al. \cite{kerola2014spectral} & Spectral Graph Skeletons & Conc & Lowlv & Hand & \checkmark & \checkmark & \checkmark &  &  &  \\\hline
Cippitelli et al. \cite{cippitelli2016human} & Key Poses & BoW & Lowlv & Dict & & \checkmark & \checkmark & \checkmark & & \checkmark\\\hline
\end{tabular}
\end{table*}

\section{Information Modality}\label{sec:InfoViews}

Skeleton-based human representations are constructed from various features computed from raw 3D skeletal data that can be possibly acquired from various sensing technologies.
We define each type of skeleton-based features extracted from each individual sensing technique
as a \emph{modality}.
From the perspective of information modality,
3D skeleton-based human representations can be classified into four categories 
based on joint displacement, orientation, raw position, and combined information.
Existing approaches falling in each categories are summarized in detail
in Tables \ref{tab:Displacement}--\ref{tab:Combined}, respectively.

\subsection{Displacement-Based Representations}
Features extracted from displacements of skeletal joints
are widely applied in many skeleton-based representations due to the
simple structure and easy implementation.
They use information from the displacement of skeletal joints,
which can either be the displacement between different human joints within the same frame or the displacement of the same joint across different time periods.

\subsubsection{Spatial Displacement Between Joints}

Representations based on relative joint displacements compute
 spatial displacements of coordinates of human skeletal joints in 3D space,
which are acquired from the same frame at a time point.

The pairwise relative position of human skeleton joints is the most widely studied displacement feature for human representation \cite{Wang_CVPR12,Wang_PAMI14,Yang_CVPRW12,chen2013online,yao2012action,zhao2013online}.
Within the same skeleton model obtained at a time point,
for each joint $\boldsymbol{p} =(x, y, z)$ in 3D space,
the difference between
the location of joint $i$ and joint $j$ is calculated by
$\boldsymbol{p}_{ij} = \boldsymbol{p}_i - \boldsymbol{p}_j, i\neq j$.
The joint locations $\boldsymbol{p}$ are often normalized,
so that the feature is invariant to the absolute body position,
initial body orientation and body size \cite{Wang_CVPR12,Wang_PAMI14,Yang_CVPRW12}.
Chen and Koskela \cite{chen2013online} implemented a similar feature extraction
method based on pairwise relative position of skeleton joints
with normalization calculated by
${\Vert\boldsymbol{p}_i-\boldsymbol{p}_j\Vert}/{\sum_{i\neq j}{\Vert\boldsymbol{p}_i-\boldsymbol{p}_j\Vert}},i\neq j$,
which is illustrated in Fig. \ref{fig:Pairwise}.

Another group of joint displacement features extracted from the same frame for skeleton-based representation construction is based on the difference to a reference joint.
In these features,
the displacements are obtained by calculating the coordinate difference of all joints with respect to a reference joint, usually manually selected.
Given the location of a joint $(x, y, z)$ and a given reference joint $(x_c, y_c, z_c)$ in the world coordinate system,
Rahmani et al. \cite{Rahmani_WACV14} defined
the spatial joint displacement as $(\Delta x, \Delta y, \Delta z) = (x, y, z) - (x_c, y_c, z_c)$,
where the reference joint can be the skeleton centroid or a manually selected, fixed joint.
For each sequence of human skeletons representing an activity, the computed
displacements along each dimension (e.g., $\Delta x$, $\Delta y$ or $\Delta z$)
are used as features to represent humans.
Luo et al. \cite{Luo_ICCV13} applied similar position information for feature extraction.
Since the joint hip center has relatively small motions for most actions,
they used that joint as the reference.

\subsubsection{Temporal Joint Displacement}

3D human representations based on temporal joint displacements
compute the location difference across a sequence of frames acquired at different time points.
Usually, they employ both spatial and temporal information to represent people in space and time.

A widely used temporal displacement feature is implemented by comparing the joint coordinates at different time steps.
Yang and Tian \cite{Yang_CVPRW12,Yang_JVCIR13} introduced a novel feature based on the position difference of joints, called EigenJoints,
which combines three categories of features including static posture, motion, and offset features. In particular,
the joint displacement of the current frame with respect to the previous frame and initial frame is calculated.
Ellis et al. \cite{Ellis_IJCV13} introduced an algorithm to reduce latency for action recognition using a 3D skeleton-based representation that depends on spatio-temporal features computed from the information in three frames:
the current frame, the frame collected 10 time steps ago, and the frame collected 30 frames ago.
Then, the features are computed as the temporal displacement among those three frames.
Another approach to construct temporal displacement representations incorporates
the object being interacted with in each pose \cite{Wei2013}.
This approach constructs a hierarchical graph to represent positions in
3D space and motion through 1D time.
The differences of joint coordinates in two successive frames are defined as the features.
Hu et al. \cite{hu2015jointly} introduced the joint heterogeneous features learning (JOULE) model through extracting the pose dynamics using skeleton data from a sequence of depth images.
A real-time skeleton tracker is used to extract the trajectories of human joints.
Then relative positions of each trajectory pair is used to construct features to distinguish different human actions.

\begin{figure}[htbp]
  \subfigure[Displacement of pairwise joints \cite{chen2013online}]{
     \label{fig:Pairwise}
    \begin{minipage}[b]{0.21\textwidth}
      \centering
        \includegraphics[width=0.6\textwidth]{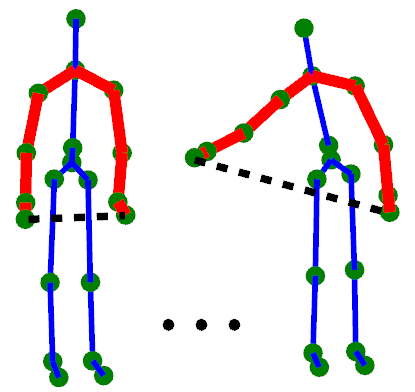}
    \end{minipage}}
  \subfigure[Relative joint displacement and joint motion volume features \cite{Rahmani_WACV14}]{
    \label{fig:MotionFeature} 
    \begin{minipage}[b]{0.22\textwidth}
      \centering
        \includegraphics[width=1.15\textwidth]{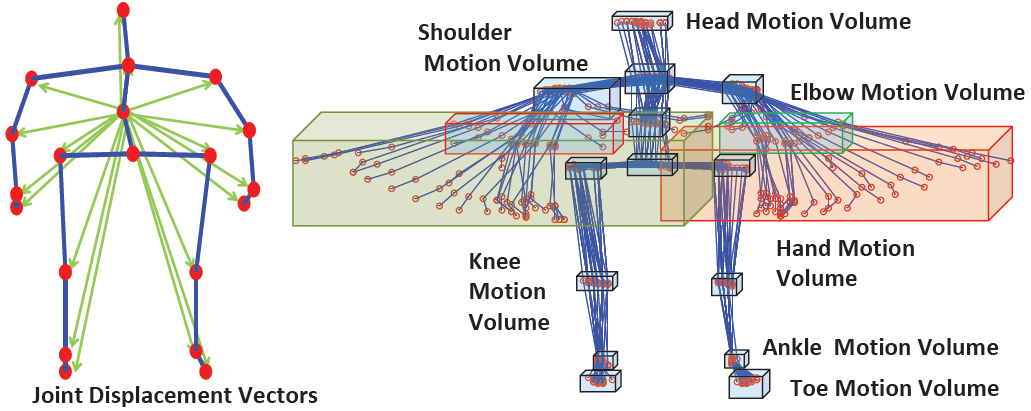}
    \end{minipage}}
  \caption{
Examples of 3D human representations based on joint displacements.
}
  \label{fig:Prediction} 
\end{figure}


The joint movement volume is another feature construction approach for human representation that also uses joint displacement information for feature extraction, especially when a joint exhibits a large movement \cite{Rahmani_WACV14}.
For a given joint,
extreme positions during the full joint motion are computed along $x$, $y$, and $z$ axes.
The maximum moving range of each joint along each dimension is then computed
by $L_{a} = \max(a_j) - \min(a_j)$, where $a=x,y,z$;
and the joint volume is defined as $V_j = L_x L_y L_z$,
as demonstrated in Fig. \ref{fig:MotionFeature}.
For each joint, $L_x , L_y , L_z$ and $V_j$ are flattened into a feature vector.
The approach also incorporates relative joint displacements with respect to the torso joint into the feature.

\begin{table*}[htbp]
\centering
\caption{Summary of 3D Skeleton-Based Representations Based on Joint Orientation Features.
Notation Is Presented in Table \ref{tab:Displacement}.
}
\label{tab:Orientation}
\small

\tabcolsep=0.1cm
\begin{tabular}{|c|c|c|c|c||c|c|c|c|c|c|}
\hline
 Reference & Approach & \tabincell{c}{Feature \\ Encoding} & \tabincell{c}{Structure \&\\ Transition} & \tabincell{c}{Feature\\ Engineering} & T & VI & ScI & SpI & OL & RT \\
\hline
\hline
Sung et al. \cite{Sung_ICRA12}\cite{sung2011human} & Orientation Matrix & Conc & Lowlv & Hand &  & \checkmark & \checkmark & \checkmark &  &   \\\hline
Xia et al. \cite{Xia_CVPRW12} & Hist. of 3D Joints & Stat & Lowlv & Hand &  & \checkmark & \checkmark &  &  & \checkmark \\\hline
Fothergill et al. \cite{MSRC12a} & Joint Angles & Conc & Lowlv & Hand & \checkmark & \checkmark & \checkmark &  & \checkmark & \checkmark \\\hline
Gu et al. \cite{gu2012human} & Gesture Recognition & BoW & Lowlv & Dict &  & \checkmark &  &  & \checkmark & \checkmark \\\hline
Jin and Choi \cite{jin2013essential} & Pairwise Orientation & Stat & Lowlv & Hand & & \checkmark & \checkmark & \checkmark & \checkmark & \checkmark \\\hline
Zhang and Tian \cite{Zhang_JCVIP12} & Pairwise Features & Stat & Lowlv & Hand &  & \checkmark &  &  & \checkmark & \checkmark \\\hline
\tabincell{c}{Kapsouras and\\ Nikolaidis \cite{kapsouras2014action}} & Dynemes Representation  & BoW & Lowlv & Dict & \checkmark &  &  &  &  &  \\\hline
\tabincell{c}{Vantigodi and\\ Radhakrishnan \cite{vantigodi2014action}} & Meta-cognitive RBF  & Stat & Lowlv & Hand & \checkmark & \checkmark & \checkmark & \checkmark &  &  \\\hline
Ohn-Bar and Trivedi \cite{Ohnbar13} & HOG2 & Conc & Lowlv & Hand & \checkmark & \checkmark & \checkmark &  &  &  \\\hline
Chaudhry et al. \cite{Chaudhry2013} & Shape from Neuroscience & BoW & Body & Dict & \checkmark & \checkmark &  &  &  &  \\\hline
Ofli et al. \cite{ofli2014sequence} & SMIJ & Conc & Lowlv & Unsup & \checkmark & \checkmark & \checkmark &  &  &  \\\hline
Miranda et al. \cite{Miranda2012} & Joint Angle & BoW & Lowlv & Dict & & \checkmark & \checkmark &  & \checkmark & \checkmark \\\hline
Fu and Santello \cite{Fu2010} & Hand Kinematics & Conc & Lowlv & Hand &  &  & & & \checkmark & \checkmark \\\hline
Zhou et al. \cite{zhou2013hierarchical} &  4D quaternions & BoW & Lowlv & Dict & \checkmark &  &  & \checkmark & \checkmark & \checkmark \\\hline
Campbell and Bobick \cite{Campbell_CVPR95} & Phase Space & Conc & Lowlv & Hand & \checkmark & \checkmark & \checkmark &  &  &  \\\hline
Boubou and Suzuki \cite{Boubou2014} & HOVV  & Stat & Lowlv & Hand & \checkmark & \checkmark & \checkmark & \checkmark &  & \checkmark \\\hline
Sharaf et al. \cite{sharaf2015real} & Joint angles and velocities & Stat & Lowlv & Hand &  \checkmark & \checkmark & \checkmark & \checkmark & \checkmark & \checkmark\\\hline
\tabincell{c}{Parameswaran\\ and Chellappa \cite{parameswaran2006view}} & ISTs & Conc & Lowlv & Hand & \checkmark & \checkmark & \checkmark & \checkmark &  &  \\\hline
\end{tabular}
\end{table*}

\subsection{Orientation-Based Representations}

Another widely used information modality for human representation construction is based on joint orientations,
since in general orientation-based features are invariant to human position, body size, and orientation to the camera.

\subsubsection{Spatial Orientation of Pairwise Joints}

Approaches based on spatial orientations of pairwise joints
compute the orientation of displacement vectors of a pair of human skeletal
joints acquired at the same time step.

A popular orientation-based human representation computes the orientation of each joint to the human centroid in 3D space.
For example, Gu et al. \cite{gu2012human} collected the skeleton data with fifteen joints and extracted features representing joint angles with respect to the
person's torso.
Sung et al. \cite{Sung_ICRA12} computed the orientation matrix of each human joint
with respect to the camera, and then transformed the joint
rotation matrix to obtain the joint orientation with respect to the person's
torso.
A similar approach was also introduced in \cite{sung2011human} based on the orientation matrix.
Xia et al. \cite{Xia_CVPRW12} introduced Histograms of 3D Joint Locations
(HOJ3D) features by assigning 3D joint positions into cone bins in 3D space.
Twelve key joints are selected and their orientation are computed with respect to the center torso point.
Using linear discriminant analysis (LDA), the features are reprojected to extract the dominant ones.
Since the spherical coordinate system used in \cite{Xia_CVPRW12} is oriented with the $x$ axis aligned with the direction a person is facing, their approach is view invariant.

Another approach is to calculate the orientation of two joints, called
relative joint orientations.
Jin and Choi \cite{jin2013essential} utilized vector orientations from one joint to another joint, named the first order orientation vector, to construct 3D human representations.
The approach also proposed a second order neighborhood that connects adjacent vectors.
The authors used a uniform quantization method to convert the continuous orientations into eight discrete
symbols to guarantee robustness to noise.
Zhang and Tian \cite{Zhang_JCVIP12} used a two mode 3D skeleton representation, combining structural data with motion data.
The structural data is represented by pairwise features, relating the
positions of each pair of joints relative to each other.
The orientation between two joints $i$ and $j$ was also used, which is given by
$
\theta(i,j)={\text{arcsin}\left(\frac{i_x-j_x}{dist(i,j)}\right)}/{2\pi},
$
where $dist(i,j)$ denotes the geometry distance between two joints $i$ and $j$ in 3D space.

\subsubsection{Temporal Joint Orientation}
Human representations based on temporal joint orientations usually compute the difference between orientations of the same joint across a temporal sequence of frames.
Campbell and Bobick \cite{Campbell_CVPR95} introduced a mapping from
the Cartesian space to the ``phase space''.
By modeling the joint trajectory in the new space,
the approach is able to represent a curve that can be easily visualized and
quantifiably compared to other motion curves.
Boubou and Suzuki \cite{Boubou2014} described a representation based on the so-called Histogram of Oriented Velocity Vectors (HOVV),
which is a histogram of the velocity orientations computed from 19 human joints
in a skeleton kinematic model acquired from the Kinect v1 camera.
Each temporal displacement vector is described by its orientation in 3D space
as the joint moves from the previous position to the current location.
By using a static skeleton prior to deal with static poses
with little or no movement,
this method is able to effectively represent humans with still poses in 3D space in human action recognition applications.

\begin{table*}[tbp]
\centering
\caption{Summary of Representations Based on Raw Position Information.
Notation Is Presented in Table \ref{tab:Displacement}.
}
\label{tab:Position}
\small

\tabcolsep=0.1cm
\begin{tabular}{|c|c|c|c|c||c|c|c|c|c|c|}
\hline
Reference & Approach & \tabincell{c}{Feature \\ Encoding} & \tabincell{c}{Structure \&\\ Transition} & \tabincell{c}{Feature\\ Engineering} & T & VI & ScI & SpI & OL & RT \\
\hline
\hline
Du et al. \cite{du2015hierarchical} & BRNNs & Conc & Body & Deep & \checkmark &  &  &  &  &  \\\hline
Paiement et al. \cite{paiementonline} & Normalized Joints & Conc & Manif & Hand & \checkmark & \checkmark & \checkmark & \checkmark &  \checkmark & \checkmark\\\hline
Kazemi et al. \cite{kazemi2013multi} & Joint Positions & Conc & Lowlv & Hand & & \checkmark & & & & \\
\hline
Seidenari et al. \cite{Seidenari2013} & Multi-Part Bag of Poses & BoW & Lowlv & Dict & \checkmark & \checkmark & \checkmark &  &  &  \\\hline
Chaaraoui et al. \cite{chaaraoui2014evolutionary} & Evolutionary Joint Selection & BoW & Lowlv & Dict &  & \checkmark &  &  &  &  \\\hline
Reyes et al. \cite{reyes2011featureweighting} & Vector of Joints & Conc & Lowlv & Hand &  & \checkmark &  & \checkmark &  &  \\\hline
Patsadu et al. \cite{patsadu2012human} & Vector of Joints & Conc & Lowlv & Hand &  & & \checkmark & \checkmark & & \\\hline
Huang and Kitani \cite{huang2014action} & Cost Topology & Stat & Lowlv & Hand &  & & & & & \\\hline
Devanne et al. \cite{devanne2015combined} & Motion Units & Conc & Manif & Hand &  & \checkmark &  &  &  &  \\\hline
Wang et al. \cite{wang2013approach} & Motion Poselets & BoW & Body & Dict &  & \checkmark &  &  &  &  \\\hline
Wei et al. \cite{wei2013concurrent} & Structural Prediction & Conc & Lowlv & Hand &  & \checkmark &  & \checkmark &  &  \\\hline
Gupta et al. \cite{gupta20143d} & 3D Pose w/o Body Parts & Conc & Lowlv & Hand &  & \checkmark &  & \checkmark &  &  \\\hline
Amor et al. \cite{boulbaba2015action} & Skeleton's Shape & Conc & Manif & Hand &  & \checkmark & \checkmark & \checkmark & &\\\hline
Sheikh et al. \cite{sheikh2005exploring} & Action Space & Conc & Lowlv & Hand & \checkmark & \checkmark & \checkmark & \checkmark &  &  \\\hline
Yilma and Shah \cite{yilma2005recognizing} & Multiview Geometry & Conc & Lowlv & Hand & \checkmark & \checkmark &  &  &  &  \\\hline
Gong et al. \cite{gong2014structured}  & Structured Time & Conc & Manif & Hand & \checkmark & \checkmark &  & \checkmark &  &  \\\hline
Rahmani and Mian \cite{rahmani2015learning} & Knowledge Transfer & BoW & Lowlv & Dict &  & \checkmark &  &    &  &  \\\hline
Munsell et al. \cite{munsell2012person} & Motion Biometrics & Stat & Lowlv & Hand & \checkmark & \checkmark &  &  &  &  \\\hline
Lillo et al. \cite{Lillo2014} & Composable Activities & BoW & Lowlv & Dict & \checkmark & \checkmark & \checkmark &  &  &  \\\hline
Wu et al. \cite{wu2015watch} & Watch-n-Patch & BoW & Lowlv & Dict & \checkmark & \checkmark &  &  & \checkmark & \checkmark \\\hline
Gong and Medioni \cite{Gong2011} & Dynamic Manifolds & BoW & Manif & Dict & \checkmark & \checkmark &  & \checkmark &  &  \\\hline
Han et al. \cite{Han2010} & Hierarchical Manifolds & BoW & Manif & Dict & \checkmark & \checkmark & \checkmark & \checkmark &  &  \\\hline
Slama et al. \cite{slama2014,slama2014grassmannian} & Grassmann Manifolds & BoW & Manif & Dict & \checkmark & \checkmark &  & \checkmark & \checkmark & \checkmark \\\hline
Devanne et al. \cite{devanne20143} & Riemannian Manifolds & Conc & Manif & Hand & \checkmark & \checkmark & \checkmark & \checkmark & \checkmark & \checkmark \\\hline
Huang et al. \cite{huang2014human} & Shape Tracking & Conc & Lowlv & Hand & \checkmark & \checkmark & \checkmark &  & \checkmark & \checkmark \\\hline
Devanne et al. \cite{devanne2013space} & Riemannian Manifolds & Conc & Manif & Hand & \checkmark & \checkmark & \checkmark & \checkmark &  &  \\\hline
Zhu et al. \cite{zhu2016cooccurrence} & RNN with LSTM & Conc & Lowlv & Deep & \checkmark &  &  &  &  &  \\\hline
Chen et al. \cite{chen2014action} & EnwMi Learning & BoW & Lowlv & Dict & \checkmark & \checkmark & \checkmark &  &  &  \\\hline
Hussein et al. \cite{Hussein2013} & Covariance of 3D Joints & Stat & Lowlv & Hand & \checkmark & \checkmark & \checkmark & \checkmark &  &  \\\hline
Shahroudy et al. \cite{shahroudy2015multimodal} & MMMP & BoW & Body & Unsup & \checkmark & \checkmark & \checkmark & & & \\\hline
Jung and Hong \cite{jung2015enhanced} & Elementary Moving Pose & BoW & Lowlv & Dict & \checkmark & \checkmark & \checkmark & \checkmark &  &  \\\hline
Evangelidis et al. \cite{evangelidis2014skeletal} & Skeletal Quad & Conc & Lowlv & Hand & \checkmark & \checkmark & \checkmark &  &  &  \\\hline
Azary and Savakis \cite{azary2013grassmannian} & Grassmann Manifolds & Conc & Manif & Hand & \checkmark & \checkmark & \checkmark & \checkmark &  &  \\\hline
Barnachon et al. \cite{barnachon2014ongoing} & Hist. of Action Poses & Stat & Lowlv & Hand &  &  &  &  & \checkmark & \checkmark \\\hline
Shahroudy et al. \cite{shahroudy2014multi} & Feature Fusion & BoW & Body & Unsup & & \checkmark & \checkmark &  &  &  \\\hline
Cavazza et al. \cite{cavazza2016kernelized} & Kernelized-COV & Stat & Lowlv & Hand & \checkmark & \checkmark & \checkmark & \checkmark &  &  \\\hline
\end{tabular}
\end{table*}

\subsection{Representations Based on Raw Joint Positions}

Besides joint displacements and orientations,
raw joint positions directly obtained from sensors are also used by many methods to construct space-time 3D human representations.

\begin{figure}[htbp]
\centering
\includegraphics[width=0.42\textwidth]{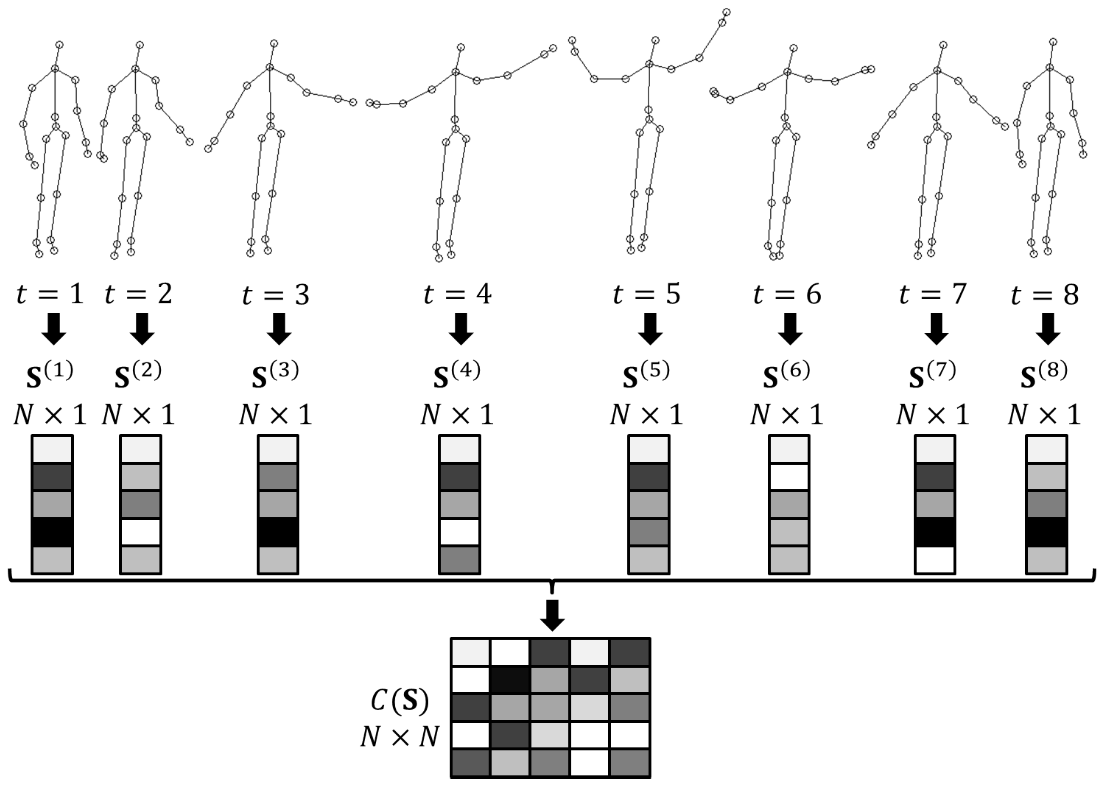}
\caption{3D human representation based on the Cov3DJ descriptor \cite{Hussein2013}.}
\label{fig:covarianceOf3dJoints}
\end{figure}

A category of approaches flatten joint positions acquired in the same frame into a column vector.
Given a sequence of skeleton frames, a matrix can be formed to naively encode the sequence with each column containing the flattened joint coordinates obtained at a specific time point.
Following this direction,
Hussein et al. \cite{Hussein2013} computed the statistical Covariance of 3D Joints (Cov3DJ) as their features,
as illustrated in Fig. \ref{fig:covarianceOf3dJoints}.
Specifically, given $K$ human joints with each joint denoted by
$\boldsymbol{p}_i = (x_i, y_i, z_i), i = 1,\dots,K$,
a feature vector is formed to encode the skeleton acquired at time $t$:
$\boldsymbol{S}^{(t)} = [x_1^{(t)}, \dots, x_K^{(t)}, y_1^{(t)},
\dots, y_K^{(t)}, z_1^{(t)}, \dots, z_K^{(t)}]^\top$.
Given a temporal sequence of $T$ skeleton frames,
the Cov3DJ feature is computed by
$C(\boldsymbol{S}) \!=\! \frac{1}{T-1}\sum_{t=1}^{T}
(\boldsymbol{S}^{(t)}\!-\!\bar{\boldsymbol{S}}^{(t)})(\boldsymbol{S}^{(t)}\!-\!\bar{\boldsymbol{S}}^{(t)})^\top$,
where $\bar{\boldsymbol{S}}$ is the mean of all $\boldsymbol{S}$.
Since not all the joints are equally informative,
several methods were proposed to select key joints that are more descriptive
\cite{chaaraoui2014evolutionary,reyes2011featureweighting,patsadu2012human,huang2014action}.
Chaaraoui et al. \cite{chaaraoui2014evolutionary} introduced an evolutionary algorithm to
select a subset of skeleton joints to form features.
Then a normalizing process was used to achieve position, scale and rotation invariance.
Similarly,
Reyes et al. \cite{reyes2011featureweighting} selected 14 joints in 3D human skeleton models without normalization for feature extraction in gesture recognition applications.

\begin{figure}[htbp]
\centering
\includegraphics[width=0.45\textwidth]{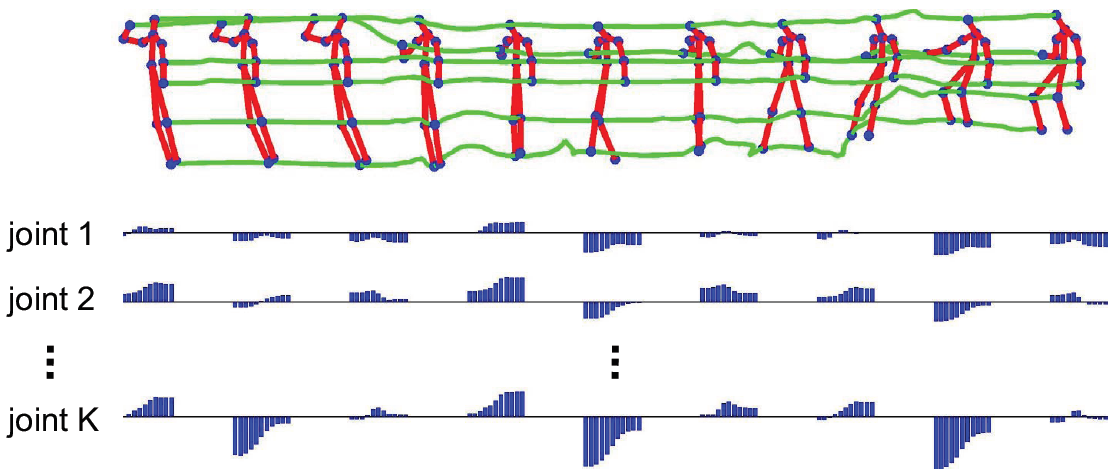}
\caption{Trajectory-based representation based on wavelet features \cite{wei2013concurrent}.}
\label{fig:WaveletICCV13Wei}
\end{figure}

Another group of representation construction techniques utilize the raw joint position information to form a trajectory, and then extract features from this trajectory,
which are often called the trajectory-based representation.
For example, Wei et al. \cite{wei2013concurrent} used a sequence of 3D human skeletal joints to construct joint trajectories,
and applied wavelets to encode each temporal joint sequence into features,
which is demonstrated in Fig. \ref{fig:WaveletICCV13Wei}.
Gupta et al. \cite{gupta20143d} proposed
a cross-view human representation,
which matches trajectory features of videos to MoCap joint trajectories
and uses these matches to generate multiple motion projections as features.
Junejo et al. \cite{junejo2011view} used trajectory-based self-similarity matrices (SSMs) to encode humans observed from different views.
This method showed great cross-view stability to represent humans in 3D space using MoCap data.

Similar to the application of deep learning techniques to extract features from images where raw pixels are typically used as input,
skeleton-based human representations built by deep learning methods generally rely on raw joint position information.
For example,
Du et al. \cite{du2015hierarchical} proposed an end-to-end hierarchical recurrent neural network (RNN) for the skeleton-based representation construction,
in which the raw positions of human joints are directly used as the input to the RNN.
Zhu et al. \cite{zhu2016cooccurrence} used raw 3D joint coordinates as the input to a RNN with Long Short-Term Memory (LSTM) to automatically learn human representations.

\subsection{Multi-Modal Representations}

Since multiple information modalities are available,
an intuitive way to improve the descriptive power of a human representation
is to integrate multiple information sources and build a multi-modal  representation to encode humans in 3D space.
For example,
the spatial joint displacement and orientation can be integrated together to build human representations.
Guerra-Filho and Aloimonos \cite{guerra2006understanding} proposed a method that
maps 3D skeletal joints to 2D points in the projection plane of the camera
and computes joint displacements and orientations of the 2D joints in the projected plane.
%
Gowayyed et al. \cite{Gowayyed_IJCAI13} developed the histogram of oriented displacements (HOD) representation
that computes the orientation of temporal joint displacement vectors and uses
their magnitude as the weight to update the histogram in order to make the representation speed-invariant.

\begin{table*}[htbp]
\centering
\caption{Summary of Representations Based on Multi-Modal Information.
Notation Is Presented in Table \ref{tab:Displacement}.
}
\label{tab:Combined}
\small

\tabcolsep=0.1cm
\begin{tabular}{|c|c|c|c|c|c||c|c|c|c|c|}
\hline
Reference & Approach & \tabincell{c}{Feature \\ Encoding} & \tabincell{c}{Structure \&\\ Transition} & \tabincell{c}{Feature\\ Engineering}  & T & VI & ScI & SpI & OL & RT \\
\hline\hline
Han et al. \cite{han2017simultaneous} & FABL & Conc & Body & Unsup & \checkmark & \checkmark & \checkmark & & & \checkmark \\\hline
Ganapathi et al. \cite{Ganapathi2010} & Kinematic Chain & Conc & Lowlv & Hand & & & & &  & \checkmark \\\hline
Ionescu et al. \cite{ionescu2014human3} & MPJPE \& MPJAE & Conc & Lowlv & Hand & \checkmark & \checkmark & \checkmark & & & \checkmark\\\hline
Marinoiu et al. \cite{marinoiu2013pictorial} &  Visual Fixation Pattern & Conc & Lowlv & Hand & & & & & &  \\\hline
Sigal et al. \cite{sigal2010humaneva} &  Parametrization of the Skeleton & Conc & Lowlv & Hand & & \checkmark & & & &  \\\hline
Huang et al. \cite{huang2014sequential} & SMMED & Conc & Lowlv & Hand & \checkmark & & & &  \checkmark & \checkmark\\\hline
Bloom et al. \cite{bloom2014g3di} & Pose Based Features & Conc & Lowlv & Hand & \checkmark & & & &  \checkmark & \checkmark\\\hline
Yu et al. \cite{yu2015discriminative} & Orderlets & Conc & Lowlv & Hand  & \checkmark & & & & \checkmark & \checkmark \\\hline
Koppula and Saxena \cite{koppula2013learning} & Node Feature Map & Conc & Lowlv & Hand & \checkmark & & & &  \checkmark & \checkmark\\\hline
Sadeghipour et al. \cite{sadeghipour2012gesture} & Spatial Positions \& Directions & Conc & Lowlv & Hand &  &  &  & &  \checkmark & \\\hline
Bloom et al. \cite{Bloom_CVPRW12} & Dynamic Features & Conc & Lowlv & Hand & \checkmark  &  &  &  & \checkmark & \checkmark \\\hline
Tenorth et al. \cite{tenorth2009tum} & Set of Nominal Features & Conc & Lowlv & Hand & &  &  & &  & \\\hline
\tabincell{c}{Guerra-Filho and\\ Aloimonos \cite{guerra2006understanding}} & Visuo-motor Primitives  & Conc & Lowlv & Hand &  & \checkmark & \checkmark & \checkmark &  &  \\\hline
Gowayyed et al. \cite{Gowayyed_IJCAI13} & HOD & Stat & Lowlv & Hand  & \checkmark & \checkmark & \checkmark & \checkmark &  &  \\\hline
Zanfir et al. \cite{Zanfir2013ICCV} & Moving Pose & BoW & Lowlv & Dict & \checkmark & \checkmark &  &  &  & \checkmark \\\hline
Bloom et al. \cite{Bloom2013} & Dynamic Features & Conc & Lowlv & Hand & \checkmark  &  &  &  & \checkmark & \checkmark \\\hline
Vemulapalli et al. \cite{Vemulapalli2014} & Lie Group Manifold & Conc & Manif & Hand & \checkmark & \checkmark & \checkmark & \checkmark &  &  \\\hline
Zhang and Parker \cite{Zhang15} & BIPOD & Stat & Body & Hand & \checkmark & \checkmark & \checkmark &  & \checkmark & \checkmark \\\hline
Lv and Nevatia \cite{lv2006recognition} & HMM/Adaboost & Conc & Lowlv & Hand & \checkmark & \checkmark & \checkmark &  &  &  \\\hline
Herda et al. \cite{herda2005hierarchical} & Quaternions & Conc & Body & Hand & & \checkmark & \checkmark &  & \checkmark & \checkmark \\\hline
Negin et al. \cite{Negin13} & RDF Kinematic Features & Conc & Lowlv & Unsup & \checkmark & \checkmark & \checkmark &  &  &  \\\hline
Masood et al. \cite{Masood2011} & Logistic Regression & Conc & Lowlv & Hand & \checkmark & & & & \checkmark & \checkmark \\\hline
Meshry et al. \cite{Meshry2016linear} & Angle \& Moving Pose & BoW & Lowlv & Unsup & \checkmark & \checkmark & & & \checkmark & \checkmark \\\hline
Tao and Vidal \cite{tao2015moving} & Moving Poselets & BoW & Body & Dict & \checkmark &  &  & &  & \\\hline
Eweiwi et al. \cite{eweiwi2015efficient} & Discriminative Action Features & Conc & Lowlv & Unsup & \checkmark & \checkmark & \checkmark &  &   & \\\hline
Wang et al. \cite{wang2015beyond} & Ker-RP & Stat & Lowlv & Hand & \checkmark & \checkmark & & & & \\\hline
Salakhutdinov et al. \cite{salakhutdinov2013learning} & HD Models & Conc & Lowlv & Deep & \checkmark & \checkmark & \checkmark & \checkmark &  & \\\hline
\end{tabular}
\end{table*}

Multi-modal space-time human representations were also actively studied,
which are able to integrate both spatial and temporal information and represent human motions in 3D space.
Yu et al. \cite{yu2015discriminative} integrated three types of features to construct a spatio-temporal representation, including pairwise joint distances, spatial joint coordinates, and temporal variations of joint locations.
Masood et al. \cite{Masood2011} implemented a similar representation by incorporating both pairwise joint distances and temporal joint location variations.
Zanfir et al. \cite{Zanfir2013ICCV} introduced the so-called moving pose feature
that integrates raw 3D joint positions as well as first and second derivatives of the joint trajectories,
based on the assumption that the speed and acceleration of human joint motions can be described accurately by quadratic functions.
%

\subsection{Summary}
Through computing the difference of skeletal joint positions in 3D real-world space,
displacement-based representations
are invariant to absolute locations and orientations of people with respect to the camera,
which can provide the benefit of forming view-invariant spatio-temporal human representations.
Similarly, orientation-based human representations can provide the same view-invariance
because they are also based on the relative information between human joints.
In addition, since orientation-based representations
do not rely on the displacement magnitude,
they are usually invariant to human scale variations.
Representations based directly on raw joint positions
are widely used due to the simple acquisition from sensors.
Although normalization procedures can make human representations
partially invariant to view and scale variations,
more sophisticated construction techniques (e.g., deep learning)
are typically needed to develop robust human representations.


Representations without involving temporal information are suitable to address problems such as pose and gesture recognition.
However, if we want the representations to be capable of
encoding dynamic human motions,
temporal information needs to be integrated.
Activity recognition can benefit from
spatio-temporal representations
that incorporate time and space information simultaneously.
Among space-time human representations,
approaches based on joint trajectories can be designed to be insensitive to motion speed invariance.
In addition,
fusion of multiple feature modalities typically results in improved performance (further analysis is provided in Section \ref{sec:performance_analysis}).
%

\section{Representation Encoding}\label{sec:RepEncoding}

Feature encoding is a necessary and important component in representation construction \cite{huang2014feature},
which aims at integrating all extracted features together into a final feature vector that can be used as the input to classifiers or other reasoning systems.
In the scenario of 3D skeleton-based representation construction,
the encoding methods can be broadly
grouped into three classes:
concatenation-based encoding,
statistics-based encoding,
and bag-of-words encoding.
The encoding technique used by each reviewed human representation is
summarized in the \emph{Feature Encoding} column in Tables \ref{tab:Displacement}--\ref{tab:Combined}.


\subsection{Concatenation-Based Approach}

We loosely define feature concatenation as a representation encoding approach,
which is a popular method to integrate multiple features into a single feature vector during human representation construction.
Many methods directly use extracted skeleton-based features,
such as displacements and orientations of 3D human joints,
and concatenate them into a 1D feature vector
to build a human representation
 \cite{reyes2011featureweighting, patsadu2012human,
Yang_CVPRW12, Yang_JVCIR13, chen2013online, Wang_PAMI14, Wei2013,
Campbell_CVPR95, sheikh2005exploring, yilma2005recognizing, gong2014structured,
Sung_ICRA12, Ohnbar13, MSRC12a, lv2006recognition,yu2015discriminative}.
For example,
Fothergill et al. \cite{MSRC12a} encoded the feature vector by concatenating 35 skeletal joint angles, 35 joint angle velocities, and 60 joint velocities into a 130-dimensional vector at each frame.
Then, feature vectors from a sequence of frames are further concatenated into a big final feature vector that is fed into a classifier for reasoning.
Similarly, Gong et al. \cite{gong2014structured} directly
concatenated 3D joint positions into a 1D vector as a representation at each frame to address the time series segmentation problem.

\subsection{Statistics-Based Encoding}
Statistics-based encoding is a common but effective method to incorporate all features into a final feature vector,
without applying any feature quantization procedure.
This encoding methodology processes and organizes features through simple statistics.
For example,
the Cov3DJ representation \cite{Hussein2013}, as illustrated in Fig. \ref{fig:covarianceOf3dJoints},
computes the covariance of a set of 3D joint position vectors collected across a sequence of skeleton frames.
Since a covariance matrix is symmetric,
only upper triangle values are utilized to form the final feature in \cite{Hussein2013}.
An advantage of this statistics-based encoding approach is that
the size of the final feature vector is independent of the number of frames.
Moreover, Wang et al. \cite{wang2015beyond} proposed an open framework by using the kernel matrix over feature dimensions as a generic representation and elevated the covariance representation to the unlimited opportunities.


The most widely used statistics-based encoding methodology is histogram encoding,
which uses a 1D histogram to estimate the distribution of extracted skeleton-based features.
For example, Xia et al. \cite{Xia_CVPRW12} partitioned the 3D space into
a number of bins using a modified spherical coordinate system
and counted the number of joints falling in each bin to form a 1D histogram,
which is called the Histogram of 3D Joint Positions (HOJ3D).
A large number of skeleton-based human representations
using similar histogram encoding methods
were also introduced,
 including
 Histogram of Joint Position Differences (HJPD)\cite{Rahmani_WACV14},
Histogram of Oriented Velocity Vectors (HOVV)\cite{Boubou2014},
and Histogram of Oriented Displacements (HOD)\cite{Gowayyed_IJCAI13},
among others
\cite{munsell2012person,vantigodi2014action,barnachon2014ongoing,huang2014action,Zhang_JCVIP12,Zhang15}.
When multi-modal skeleton-based features are involved, concatenation-based encoding is usually employed
to incorporate multiple histograms into a single final feature vector \cite{Zhang15}.


\subsection{Bag-of-Words Encoding} \label{sec:sub:DictEncode}
Unlike concatenation and statistics-based encoding methodologies,
bag-of-words encoding applies a coding operator
to project each high-dimensional feature vector into a single code (or word) using a learned codebook (or dictionary) that contains all possible codes.
This procedure is also referred to as feature quantization.
Given a new instance,
this encoding methodology uses the normalized frequency vector of code occurrence as the final feature vector.
Bag-of-words encoding is widely employed by a large number of skeleton-based human representations
\cite{chaaraoui2014evolutionary,Luo_ICCV13,Jiang2014,Zanfir2013ICCV,rahmani2015learning,hu2015jointly,Zou13,Seidenari2013,Lillo2014,wu2015watch,wang2014cross,xiaohan2015joint,Gong2011,slama2014,gu2012human,zhao2013online,Chaudhry2013,Miranda2012,wang2013approach,kapsouras2014action,Meshry2016linear,tao2015moving}.
According to how the dictionary is learned,
the encoding methods can be broadly categorized into two groups,
based on clustering or sparse coding.

The k-means algorithm is a popular unsupervised learning method that is commonly used to construct a dictionary.
Wang et al. \cite{wang2013approach} grouped human joints into five body parts,
and used the k-means algorithm to cluster the training data.
The indices of the cluster centroids are utilized as codes to form a dictionary.
During testing, query body part poses are quantized using the learned dictionary.
Similarly, Kapsouras and Nikolaidis\cite{kapsouras2014action}
used the k-means clustering method on skeleton-based features
consisting of joint orientations
and orientation differences in multiple temporal scales,
in order to select representative patterns to build a dictionary.

\begin{figure}[htbp]
\centering
\includegraphics[width = 0.45\textwidth]{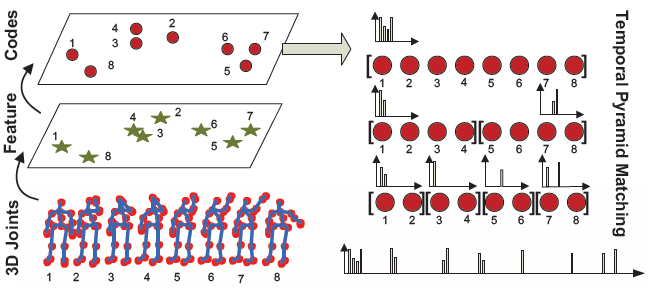}
\caption{Dictionary learning based on sparse coding for skeleton-based human representation construction \cite{Luo_ICCV13}.}\label{fig:ICCV13Luo}
\end{figure}

Sparse coding is another common approach to construct efficient representations of data as a (often linear) combination of a set of distinctive patterns (i.e., codes) learned from the data itself.
Zhao et al. \cite{zhao2013online} introduced a sparse coding approach regularized by the $l_{2,1}$ norm to construct a dictionary of templates from the so-called Structured Streaming Skeletons (SSS) features in a gesture recognition application.
Luo et al. \cite{Luo_ICCV13} proposed another sparse coding method to learn
 a dictionary based on pairwise joint displacement features.
This approach uses a combination of group sparsity and geometric constraints to select sparse and more representative patterns as codes.
An illustration of the dictionary learning method to encode skeleton-based human representations is presented in Fig. \ref{fig:ICCV13Luo}.

\subsection{Summary}

Due to its simplicity and high efficiency,
the concatenation-based feature vector construction method
is widely applied in real-time online applications to reduce processing latency.
The method is also used to integrate features from multiple sources into a single vector for further encoding/processing.
By not requiring a feature quantization process,
statistics-based encoding, especially based on histograms,
is efficient and relatively robust to noise.
However, the statistics-based encoding method
is incapable of identifying the representative patterns and modeling the structure of the data,
thus making it lacking in discriminative power.
Bag-of-words encoding can automatically find a good over-complete basis
and encode a feature vector using a sparse solution to minimize approximation error.
Bag-of-words encoding is also validated to be robust to data noise.
However, dictionary construction and feature quantization require additional computation.
According to the performance reported by the papers (as further analyzed in Section \ref{sec:performance_analysis}),
the bag-of-words encoding can generally obtain superior performance.

\section{Structure and Topological Transition}\label{sec:Hierarchy}

While most skeleton-based 3D human representations are based on pure low-level features
extracted from the skeleton data in 3D Euclidean space,
several works studied mid-level features or feature transition to other topological space.
This section categorizes the reviewed approaches from the structure and transition perspective into three groups:
representations using low-level features in Euclidean space,
representations using mid-level features based on human body parts,
and manifold-based representations.
The major class of each representation categorized from this perspective is listed in the \emph{Structure and Transition} column in Tables \ref{tab:Displacement}--\ref{tab:Combined}.

\subsection{Representations Based on Low-Level Features}

A simple, straightforward framework to construct skeleton-based representations
is to use low-level features computed from 3D skeleton data in Euclidian space,
without considering human body structures or applying feature transition.
Most of the existing representations fall in this category.
The representations can be constructed by single-layer methods,
or by approaches with multiple layers.

An example of the single-layer representation construction method is
the EigenJoints approach introduced by Yang and Tian \cite{Yang_CVPRW12,Yang_JVCIR13}.
This approach extracts low-level features from skeletal data,
such as pairwise joint displacements, and uses Principal Component Analysis (PCA) to perform  dimension reduction.
Many other existing human representations are also
based on low-level skeleton-based features
\cite{Bloom2013,Bloom_CVPRW12,Yao2012,Luo_ICCV13,Yun_CVPRW12,Sung_ICRA12,Zhang12,Zou13,Ohnbar13,Miranda2012,Fu2010,Seidenari2013,sharaf2015real}
without modeling the hierarchy of the data.

\begin{figure}[htbp]
\centering
\includegraphics[width=0.45\textwidth]{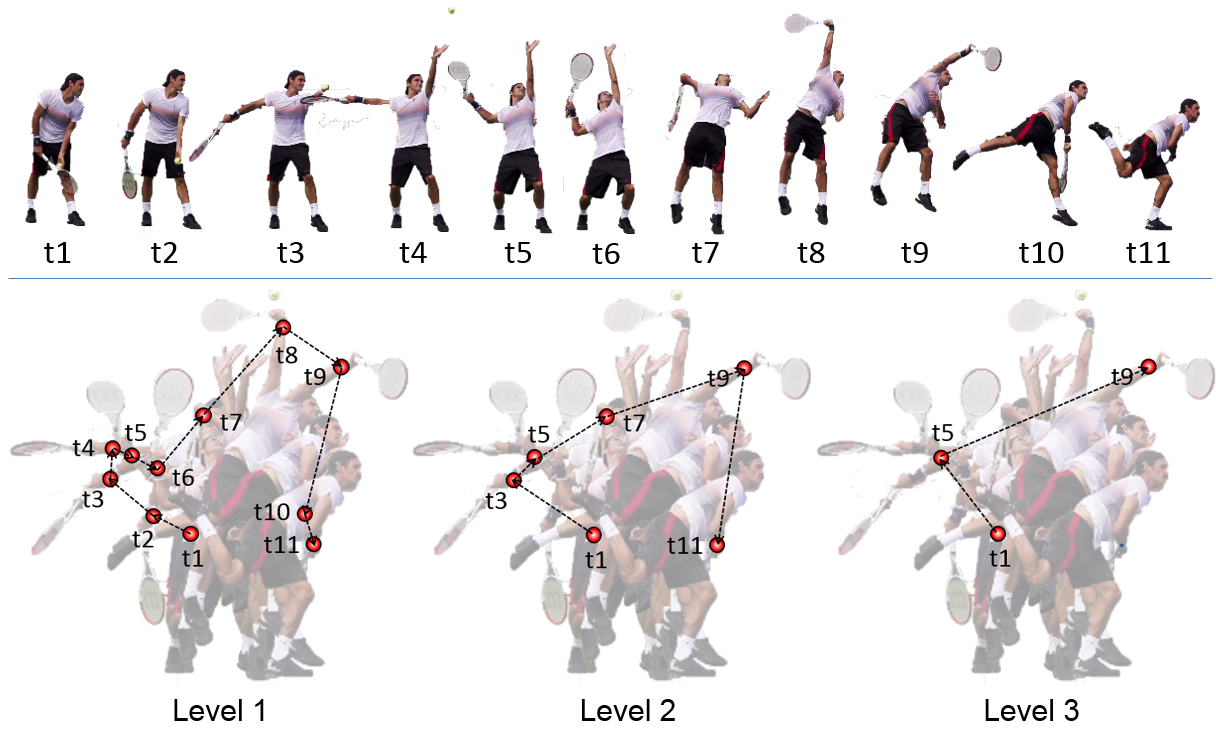}
\caption{Temporal pyramid techniques to incorporate multi-layer temporal information for space-time human representation construction based on a sequence of 3D skeleton frames \cite{Zhang15}.
}\label{fig:hierarchies}
\end{figure}

Several multi-layer techniques were also implemented to create skeleton-based human representations from low-level features.
In particular, deep learning approaches inherently consist of multiple layers
with the intermediate and output layers encoding different levels of features \cite{zeiler2014visualizing}.
The multi-layer deep learning approaches have attracted an increasing attention
in recent several years
to learn human representations directly from human joint positions
\cite{zhu2016cooccurrence,Wu2014}.
Inspired by the spatial pyramid method \cite{lazebnik2006beyond} to incorporate multi-layer image information,
temporal pyramid methods were introduced and used by several skeleton-based human representations
to capture the multi-layer information in the time dimension \cite{Wang_CVPR12, Wang_PAMI14,Gowayyed_IJCAI13,Hussein2013,Zhang15}.
For example, a temporal pyramid method was proposed by Zhang et al. \cite{Zhang15} to capture long-term dependencies,
as illustrated in Fig. \ref{fig:hierarchies}.
In this example, a temporal sequence of eleven frames is used to represent a tennis-serve motion,
and the joint of interest  is the right wrist, as denoted by the red dots in Fig \ref{fig:hierarchies}.
When three levels are used in the temporal pyramid,
level 1 uses human skeleton data at all time points ($t_1, t_2, \dots, t_{11}$);
level 2 selects the joints at odd time points ($t_1, t_3, \dots, t_{11}$);
and level 3 continues this selection process and
keeps half of the temporal data points ($t_1, t_5, t_{9}$) to compute long-term orientation changes.

\subsection{Representations Based on Body Part Models}

Mid-level features based on body part models are also used
to construct skeleton-based human representations.
Since these mid-level features partially take into account
the physical structure of human body,
they can usually result in improved discrimination power to represent humans \cite{ionescu2014iterated,Zhang15,tao2015moving}.


\begin{figure}[htbp]
\centering
\includegraphics[width=0.45\textwidth]{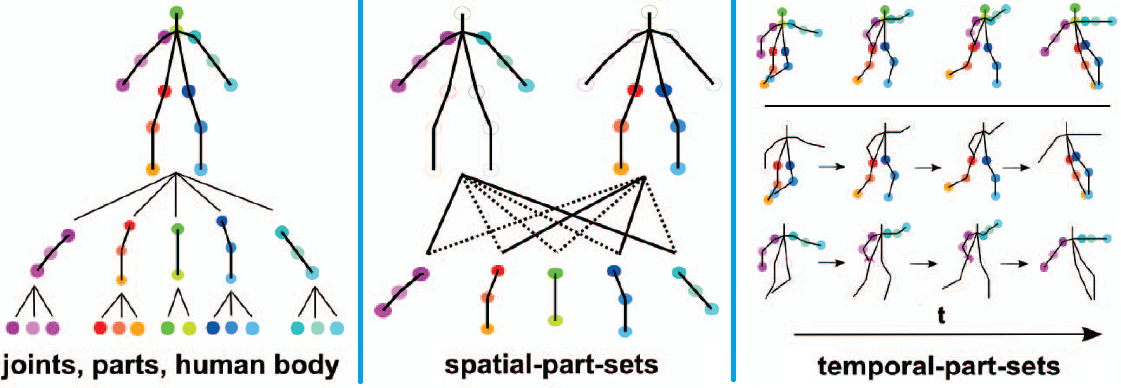}
\caption{Spatiotemporal human representations based mid-level features extracted from human body parts \cite{wang2013approach}.
}\label{fig:PoseBasedCVPR13Wang}
\end{figure}

Wang et al. \cite{wang2013approach} decomposed a kinematic human body model into five parts,
including the left/right arms/legs and torso, each consisting of a set joints.
Then, the authors used a data mining technique to obtain a spatiotemporal
human representation,
by capturing spatial configurations of body parts in one frame (by spatial-part-sets)
as well as body part movements across a sequence of frames (by temporal-part-sets),
as illustrated in Fig. \ref{fig:PoseBasedCVPR13Wang}.
With this human representation, the approach was able to obtain a hierarchical data
that can simultaneously model the correlation and motion of human joints and body parts.
Nie et al. \cite{xiaohan2015joint} implemented a spatial-temporal And-Or graph model
to represent humans at three levels including poses, spatiotemporal-parts, and parts.
The hierarchical structure of this body model captures the geometric and appearance variations of humans at each frame.
Du et al. \cite{du2015hierarchical} introduced a deep neural network to create a body part model
and investigate the correlation of body parts.

\begin{figure}[htb]
\centering
\includegraphics[width=0.45\textwidth]{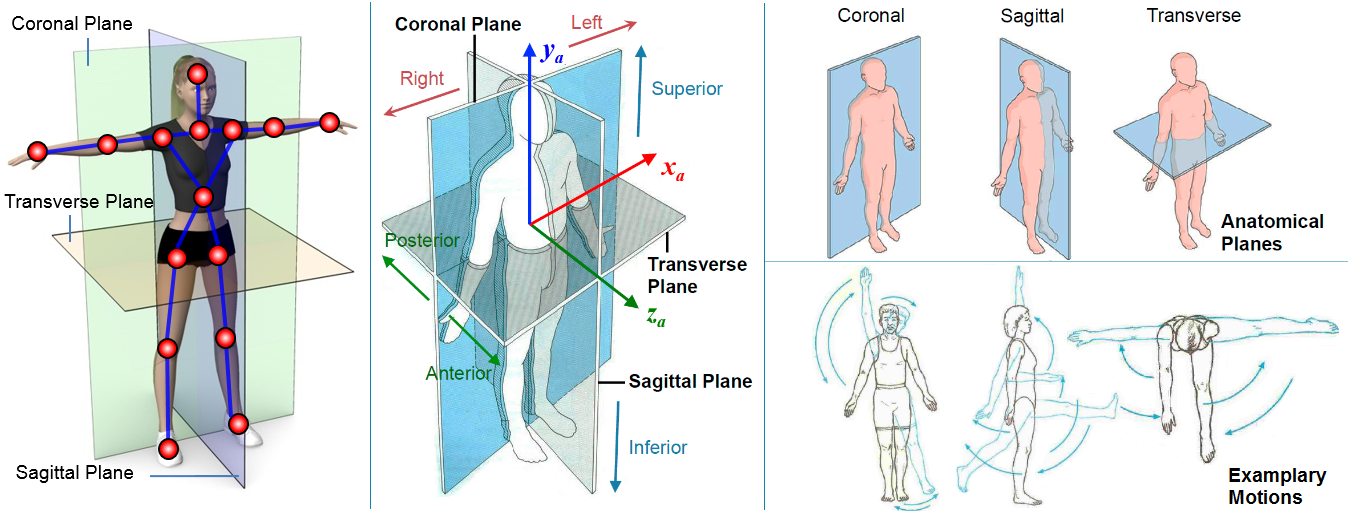}
\caption{
Representations based on mid-level features extracted from bio-inspired body part models,
inspired by human anatomy research \cite{Zhang15}.
}\label{fig: bipod}
\end{figure}


Bio-inspired body part methods were also introduced
to extract mid-level features for skeleton-based representation construction,
based on body kinematics or human anatomy.
Chaudhry et al. \cite{Chaudhry2013} implemented a bio-inspired mid-level feature
to represent people based on 3D skeleton information
through leveraging findings in the area of static shape encoding in the
neural pathway of the primate cortex \cite{Yamane2008}. 
By showing primates various 3D shapes and measuring the neural response
when changing different parameters of the shapes, the primates'
internal shape representation can be estimated,
which was then applied to extract body parts to construct skeleton-based representations.
Zhang and Parker \cite{Zhang15}
implemented a bio-inspired predictive orientation decomposition (BIPOD)
using mid-level features to construct representations of people from 3D skeleton trajectories,
which is inspired by biological research in human anatomy.
This approach decomposes a human body model into five body parts,
and then projects 3D human skeleton trajectories onto three anatomical planes
(i.e., coronal, transverse and sagittal planes),
as illustrated in Fig. \ref{fig: bipod}.
By estimating future skeleton trajectories,
the BIPOD representation possesses the ability to predict future human motions.

\subsection{Manifold-Based Representations}
A number of methods in the literature transited the skeleton data
in 3D Euclidean space to another topological space (i.e., manifold)
in order to process skeleton trajectories as curves within the new space.
This category of methods typically utilizes a trajectory-based representation.

\begin{figure}[htb]
  \subfigure[Lie group \cite{Vemulapalli2014}]{
    \label{fig:lieGroup} 
    \begin{minipage}[b]{0.22\textwidth}
      \centering
        \includegraphics[height=0.9in]{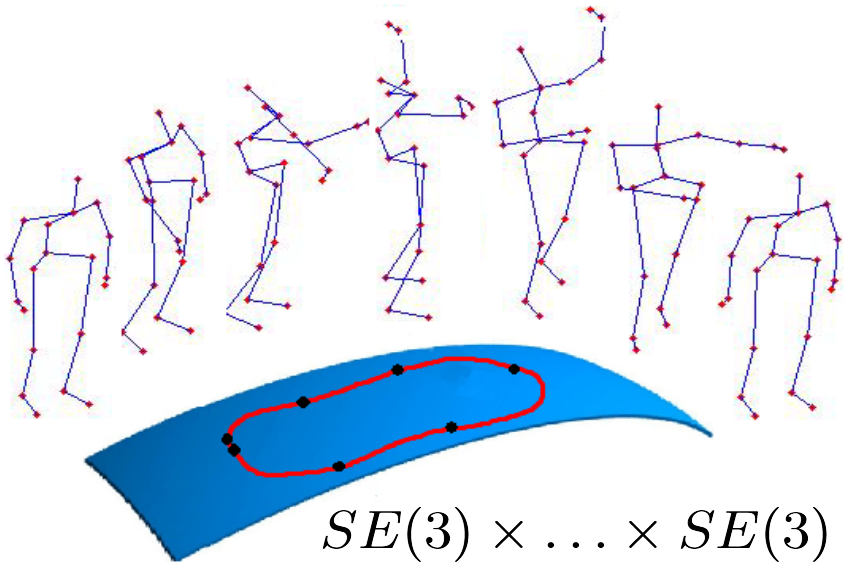}
    \end{minipage}}
  \hspace{3pt}
  \subfigure[Grassmann manifold \cite{azary2013grassmannian}]{
    \label{fig:grassmannManifold} 
    \begin{minipage}[b]{0.22\textwidth}
      \centering
        \includegraphics[height=0.9in]{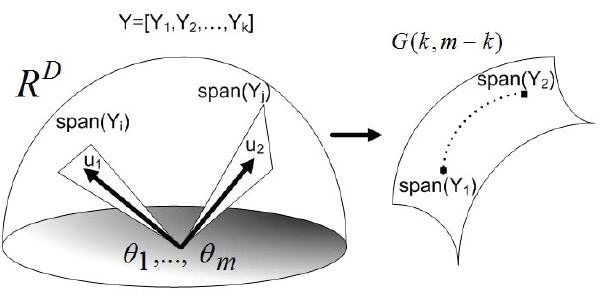}
    \end{minipage}}
  \caption{
Examples of skeleton-based representations created by transiting joint trajectories in 3D Euclidean space to a manifold.}
  \label{fig:manifold} 
\end{figure}

Vemulapalli et al. \cite{Vemulapalli2014}
introduced a skeletal representation that was created in the Lie group $SE(3)\!\times\dots\!\times\!SE(3)$,
which is a curved manifold,
based on the observation that 3D rigid body motions are members of the space.
Using this representation,
joint trajectories can be modeled as curves in the Lie group, shown in Fig. \ref{fig:lieGroup}.
This manifold-based representation can model 3D geometric relationships
between joints using rotations and translations in 3D space.
Since analyzing curves in the Lie group is not easy,
the approach maps the curves from the Lie group to its Lie algebra, which is a vector space.
Gong and Medioni \cite{Gong2011}  introduced a spatio-temporal manifold
and a dynamic manifold warping method,
which is an adaptation of dynamic time warping methods for the manifold space.
Spatial alignment is also used to deal with variations of viewpoints and body scales.
Slama et al. \cite{slama2014} introduced a multi-stage method based on a Grassmann manifold.
Body joint trajectories are represented as points on the manifold,
and clustered to find a `control tangent' defined as the mean of a cluster.
Then a query human joint trajectory is projected against the tangents to form a final representation.
This manifold was also applied by Azary and Savakis \cite{azary2013grassmannian} to build sparse human representations, shown in Fig. \ref{fig:grassmannManifold}.
Anirudh et al. \cite{anirudh2015elastic} introduced
the transport square-root velocity function (TSRVF)
to encode humans in 3D space, which provides an elastic metric to model joint trajectories on Riemannian manifolds.
Amor et al. \cite{boulbaba2015action} proposed to model the evolution of human skeleton shapes as trajectories on Kendall's shape manifolds,
and used a parameterization-invariant metric \cite{su2014rate} for
aligning, comparing, and modeling skeleton joint trajectories,
which can deal with noise caused by
large variability of execution rates within and across humans.
Devanne et al. \cite{devanne20143}
introduced a human representation by comparing the similarity between human skeletal joint trajectories in a Riemannian manifold \cite{karcher1977riemannian}.

\subsection{Summary}

Single or multi-layer human representations based on low-level features
directly extract features from 3D skeletal data without considering the physical structure of human body.
The kinematic body structure is coarsely encoded by human representations
based on mid-level features extracted from body part models,
which can capture the relationship of not only joints but also body parts.
Manifold-based representations
map motion joint trajectories into a new topological space,
in the hope of finding a more descriptive representation in the new space.
Good performance of all these human representations was reported in the literature.
However, with the complexity increment of the activities, especially long-time activities, low-level feature structures may not a good choice due to their limited representation capability.
In this case, body-part models and manifold-based representations can often improve recognition performance.

\section{Feature Engineering}\label{sec:Construction}

Feature engineering is one of the most fundamental research problems in computer vision and machine learning research.
Early feature engineering techniques for human representation construction are manual;
features are hand-crafted and their importance are manually decided.
In recent years, we have been witnessing a clear transition
from manual feature
engineering
to automated feature learning and extraction.
In this section, we categorize and analyze human representations based on 3D skeleton data from the perspective of feature engineering.
The feature engineering approach used by each human
representation is summarized in the \emph{Feature Engineering} column in Tables \ref{tab:Displacement}--\ref{tab:Combined}.

\subsection{Hand-Crafted Features}

Hand-crafted features are manually designed and constructed to capture certain geometric, statistical, morphological, or other attributes of 3D human skeleton data,
which dominated the early skeleton-based feature extraction methods
and are still intensively studied in modern research.


Lv and Nevatia \cite{lv2006recognition} decomposed the high dimensional 3D joint space
into a set of feature spaces where each of them corresponds to the
motion of a single joint or a combination of related multiple joints.
Ofli et al. \cite{ofli2014sequence} proposed a human representation
called the Sequence of the Most Informative Joints (SMIJ),
by selecting a subset of skeletal joints to extract category-dependent features.
Zhao et al. \cite{chen2013online} described a method of representing humans
using the similarity of current and previously seen skeletons in a gesture recognition application.
Pons-Moll et al. \cite{pons2014posebits} used qualitative attributes
of the 3D skeleton data, called posebits, to estimate human poses,
by manually defining features such as joint distance, articulation angle, relative position, etc.
Huang et al. \cite{huang2014human} proposed to utilize hand-crafted features
including  skeletal joint positions
to locate key frames and track humans from a multi-camera video.
In general, the majority of the existing skeleton-based human representations
employ hand-crafted features,
especially, the methodologies based on histograms and manifolds,
as presented by Tables \ref{tab:Displacement}--\ref{tab:Combined}.

\subsection{Representation Learning}

In many vision and reasoning tasks,
good performance is all about the right representation.
Thus, automated learning of skeleton-based features has become highly active
in the task of human representation construction based on 3D skeletal data.
These skeleton-based representation learning methods can be broadly divided into three groups:
dictionary learning, unsupervised feature learning, and deep learning.

\subsubsection{Dictionary Learning}

Dictionary learning aims at learning a basis set (dictionary)
to encode a feature vector as a sparse linear combination of basis elements,
as well as to adapt the dictionary to the data in a specific task.
Learning a dictionary is the foundation of the bag-of-words encoding.
In the literature of 3D skeleton-based representation creation,
the k-means algorithm \cite{wang2013approach,kapsouras2014action}
and sparse coding \cite{zhao2013online,Luo_ICCV13} are the most commonly used techniques for dictionary learning. A number of these methods are reviewed in
Section \ref{sec:sub:DictEncode}.

\subsubsection{Unsupervised Feature Learning}

The objective of unsupervised feature learning is to discover low-dimensional features that capture the underlying structure of the input data in a higher dimension.
For example, the traditional PCA method is applied for dimension reduction to extract low-dimensional features from raw skeleton features \cite{Yang_CVPRW12,Yang_JVCIR13,Meshry2016linear}.
Negin et al. \cite{Negin13} designed a feature selection method
to build human representations from 3D skeletal data.
This approach describes humans via a collection of time-series feature computed from the skeletal data,
and discriminatively optimizes a random decision forest model over this collection to identify the most effective set of features in time and space dimensions.

Very recently,
several multi-modal feature learning approaches via sparsity-inducing norms were introduced to integrate different types of features,
such as color-depth and skeleton-based features,
to produce a compact, informative 3D representation of people.
Shahroudy et al. \cite{shahroudy2014multi}
recently developed a multi-modal feature learning method to fuse the RGB-D and skeletal information into an integrated set of discriminative features.
This approach uses the group-$l_1$ norm to force features from the same view to be activated or deactivated together,
and applies the $l_{2,1}$ norm to allow a single feature within a deactivated view to be activated.
The authors also introduced a multi-modal multi-part human representation based on a hierarchical mixed norm \cite{shahroudy2015multimodal}, which regularizes
structured features of each joint subset and applies sparsity between them.
Another heterogenous feature learning algorithm
was introduced by Hu et al. \cite{hu2015jointly}.
The approach casted joint feature learning as a least-square optimization problem
that employs the Frobenius matrix norm as the regularization term that provides an efficient, closed-form solution.

\subsubsection{Deep Learning}

\begin{figure}[tbp]
\centering
\includegraphics[width=0.45\textwidth]{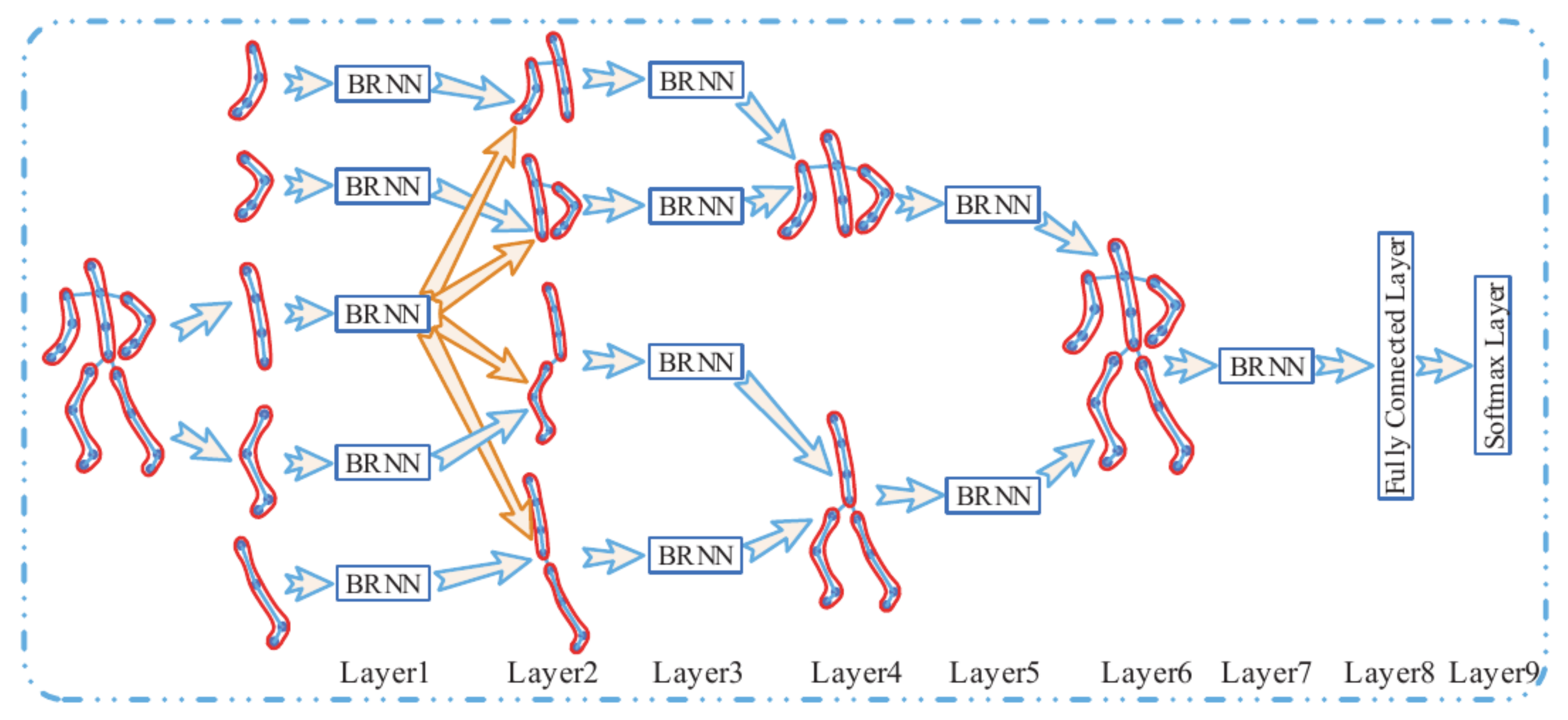}
\caption{Hierarchical RNNs for representation learning based on 3D skeletal joint locations \protect\cite{du2015hierarchical}.}
\label{fig:deepLearnedFeature}
\end{figure}

While unsupervised feature learning allows for assigning a weight to each feature element,
this methodology still relies on manually crafted features as the initial set.
Deep learning, on the other hand, attempts to automatically learn a multi-level representation directly from raw data,
by exploring a hierarchy of factors that may explain the data.
Several such approaches were developed to learn human representations from 3D skeletal joint positions directly acquired by sensors in recent several years.
For example, Du et al. \cite{du2015hierarchical} proposed an end-to-end hierarchical recurrent neural network (RNN) to construct a skeleton-based human representation.
In this method, the whole skeleton is divided into five parts according to human physical structure,
and separately fed into five bidirectional RNNs.
As the number of layers increases,
the representations extracted by the subnets are hierarchically fused to build a higher-level representation,
as illustrated in Fig. \ref{fig:deepLearnedFeature}.
Zhu et al. \cite{zhu2016cooccurrence} introduced a method based on
RNNs with Long Short-Term Memory (LSTM)
to automatically learn human representations and model long-term temporal dependencies.
In this method, joint positions are used as the input at each time slot to the LST-RNNs
that can model the joint co-occurrences to characterize human motions.
Wu and Shao \cite{Wu2014} proposed to utilize deep belief networks to model the distribution of skeleton joint locations and extract high-level features to represent humans at each frame in 3D space.
Salakhutdinov et al. \cite{salakhutdinov2013learning} proposed
a compositional learning architecture that integrates deep learning models with structured hierarchical Bayesian models.
Specifically, this approach learns a hierarchical Dirichlet process
(HDP) prior over top-level features in a deep Boltzmann machine (DBM),
which simultaneously learns low-level generic features, high-level features that capture
the correlation among the low-level features, and a category hierarchy for sharing priors over the high-level features.

\subsection{Summary}

Hand-crafted features still dominate human representations based on 3D skeletal data in the literature.
Although several approaches showed great performance various applications,
hand-crafting these manual features typically requires significant domain knowledge and careful parameter tuning. Most hand-crafted feature extraction methods are sensitive to their parameters values; poor parameter tuning can dramatically decrease the recognition performance. Also, the requirement of domain knowledge makes hand-crafted features not robust to various situations.
Unsupervised dictionary and feature learning approaches can automatically determine
which types of skeleton-based features or templates are more representative,
although they typically use hand-craft features as the input.
Deep learning, on the other hand,
can directly work with the raw skeleton information,
and automatically discover and create features.
However, deep learning methods are typically
computationally expensive,
which currently might not be suitable for online, real-time applications.

\section{Discussion}\label{sec:discussion}
\subsection{Performance Analysis of the Current State of the Art}\label{sec:performance_analysis}
In this section, we compare the accuracy and efficiency of different approaches using several most used datasets, including MSR Action3D, CAD-60, MSRC-12, and HDM05, which cover both structured light sensors (Kinect v1) and motion capture sensor systems. The performance is evaluated using the precision metric, since almost all the existing approaches report the precision results. The detailed comparison of different approaches is presented in Table \ref{tab:performance_comparison}.

\begin{table*}[htbp]
\centering
\caption{Performance Comparison among Different Approaches over Popular Datasets.}
\scriptsize
\label{tab:performance_comparison}
Notation:
In the \emph{Modality} column: \underline{D}isplacement,
\underline{O}rientation, Raw Joint \underline{P}osition, \underline{M}ulti-Modal;
In the \emph{Complexity} column: \underline{R}eal-\underline{T}ime.
\small
\tabcolsep=0.1cm
\begin{tabular}{|c|c|c||c|c|c|c||c|c|}
\hline
Dataset & Ref. & Approach & Modality & \tabincell{c}{Feature \\ Encoding} & \tabincell{c}{Structure \&\\ Transition} & \tabincell{c}{Feature\\ Engineering} & \tabincell{c}{Precision\\($\%$)} & \tabincell{c}{Complexity}\\\hline\hline
\multirow{10}{*}{MSR Action3D} & \cite{wang2015beyond} & Ker-RP & M & Stat & Lowlv & Hand & 96.9 & \\\cline{2-9}
& \cite{cavazza2016kernelized} & Kernelized-COV & P & Stat & Lowlv & Hand & 96.2 & \\\cline{2-9}
& \cite{Meshry2016linear} & Angle \& Moving Pose & M & BoW & Lowlv & Unsup & 96.1 & RT \\\cline{2-9}
& \cite{du2015hierarchical} & BRNNs & P & Conc & Body & Deep & 94.5 & \\\cline{2-9}
& \cite{chaaraoui2014evolutionary} & Joint Selection & P & BoW & Lowlv & Dict & 93.5 & \\\cline{2-9}
& \cite{shahroudy2015multimodal} & MMMP & P & BoW & Body & Unsup & 93.1 & \\\cline{2-9}
& \cite{Vemulapalli2014} & Lie Group Manifold & M & Conc & Manif & Hand & 92.5 & \\\cline{2-9}
& \cite{Zanfir2013ICCV} & Moving Pose & M & BoW & Lowlv & Dict & 91.7 & RT \\\cline{2-9}
& \cite{boulbaba2015action} & Skeleton's Shape & P & Conc & Manif & Hand & 89.0 & \\\cline{2-9}
& \cite{Wang_PAMI14} & Actionlet & D & Conc & Lowlv & Hand & 88.2 & \\\cline{2-9}
& \cite{Yang_JVCIR13} & EigenJoints & D & Conc & Lowlv & Unsup & 83.3 & RT \\\cline{2-9}
& \cite{Xia_CVPRW12} & Hist. of 3D Joints & O & Stat & Lowlv & Hand & 78.0 & RT\\\cline{2-9}
\hline
\multirow{6}{*}{CAD-60} & \cite{cippitelli2016human} & Key Poses & D & BoW & Lowlv & Dict & 93.9 & RT \\\cline{2-9}
& \cite{Zhang_JCVIP12} & Pairwise Features & O & Stat & Lowlv & Hand & 81.8 & RT \\\cline{2-9}
& \cite{koppula2013learning} & Node Feature Map & M & Conc & Lowlv & Hand & 80.8 & RT \\\cline{2-9}
& \cite{Wang_PAMI14} & Actionlet & D & Conc & Lowlv & Hand & 74.7 & \\\cline{2-9}
& \cite{Yang_JVCIR13} & EigenJoints & D & Conc & Lowlv & Unsup & 71.9 & RT \\\cline{2-9}
& \cite{sung2011human,Sung_ICRA12} & Orientation Matrix & O & Conc & Lowlv & Hand & 67.9 & \\\cline{2-9}
\hline
\multirow{6}{*}{MSRC-12} & \cite{jung2015enhanced} & Elementary Moving Pose & P & BoW & Lowlv & Dict & 96.8 & \\\cline{2-9}
& \cite{cavazza2016kernelized} & Kernelized-COV & P & Stat & Lowlv & Hand & 95.0 & \\\cline{2-9}
& \cite{wang2015beyond} & Ker-RP & M & Stat & Lowlv & Hand & 92.3 & \\\cline{2-9}
& \cite{Hussein2013} & Covariance of 3D Joints & P & Stat & Lowlv & Hand & 91.7 & \\\cline{2-9}
& \cite{Negin13} & RDF Kinematic Features & M & Conc & Lowlv & Unsup & 76.3 & \\\cline{2-9}
& \cite{zhao2013online} & Motion Templates & D & BoW & Lowlv & Dict & 66.6 & RT \\\cline{2-9}
& \cite{MSRC12a} & Joint Angles & O & Conc & Lowlv & Hand & 54.9 & RT \\\cline{2-9}
\hline
\multirow{6}{*}{HDM05} & \cite{cavazza2016kernelized} & Kernelized-COV & P & Stat & Lowlv & Hand & 98.1 & \\\cline{2-9}
& \cite{wang2015beyond} & Ker-RP & M & Stat & Lowlv & Hand & 96.8 & \\\cline{2-9}
& \cite{Zhang15} & BIPOD & M & Stat & Body & Hand & 96.7 & RT \\\cline{2-9}
& \cite{Hussein2013} & Covariance of 3D Joints & P & Stat & Lowlv & Hand & 95.4 & \\\cline{2-9}
& \cite{evangelidis2014skeletal} & Skeletal Quad & P & Conc & Lowlv & Hand & 93.9 & \\\cline{2-9}
& \cite{Chaudhry2013} & Shape from Neuroscience & O & BoW & Body & Dict & 91.7 & \\\cline{2-9}
& \cite{ofli2014sequence} & SMIJ & O & Conc & Lowlv & Unsup & 84.4 & \\\hline
\end{tabular}
\end{table*}

From Table \ref{tab:performance_comparison},
it is observed that there is no single approach that is able to guarantee the best performance over all datasets.
Performance of each approach varies when applied to different benchmark datasets. Generally, methods using multimodal information can have better activity recognition performance in comparison to methods based on single feature modality.
We can also observe that the bag-of-words feature encoding is able to improve the performance.
For feature structure $\&$ transition, the reviewed representations obtain similar recognition performance as shown in Table \ref{tab:performance_comparison}. For feature engineering, learning-based methods, including deep learning, unsupervised feature learning and dictionary learning are proved to provide superior activity recognition results in comparison to traditional hand-crafted feature engineering methods.

As a side note, several public software packages are available, which implement 3D skeletal representations of people.
The representations with open-source implementations include {Ker-RP} \cite{wang2015beyond}, {lie group manifold} \cite{Vemulapalli2014}, {orientation matrix} \cite{Sung_ICRA12}, {temporal relational features} \cite{koppula2013learningICML}, {node feature map} \cite{koppula2013learning}.
We provide the web link to these open-source packages in the reference \cite{ker-rp,lie_group_manifold,orientation_matrix,temporal_relational_features,node_feature_map,mpjpe}.

\subsection{Future Research Directions}\label{sec:future_directions}

Human representations based on 3D skeleton data can possess several desirable attributes,
including the ability to incorporate spatio-temporal information,
invariance to variations of
viewpoint, human body scale, and motion speed,
and real-time, online performance.
The characteristics of each reviewed
representation are presented in Tables \ref{tab:Displacement}--\ref{tab:Combined}.
While significant progress has been achieved on human representations based on 3D skeletal data,
there are still numerous research opportunities.
Here we briefly summarize
some of the prevalent problems and provide possible future directions.

\begin{itemize}
\item \emph{Fusing skeleton data with human texture and shape models.}
Although 3D skeleton data can be applied to construct descriptive representations of humans,
it is incapable of encoding texture information, and therefore cannot effectively represent human-object interaction.
In addition, other human models such as shape-based representation can also increase the description capability of humans.
Integrating texture and shape information with skeleton data to build a multisensory representation
has the potential to address this problem \cite{Li2010,yu2015discriminative,shahroudy2015multimodal}
and improve the descriptive power of the existing space-time human representations.

\item \emph{General representation construction via cross-training.}
A variety of devices can provide skeleton data but with different kinematic models.
It is desirable to develop cross-training methods
that can utilize skeleton data from different devices
 to build a general representation that works with different skeleton models \cite{Zhang15}.
A method of unifying skeleton data to the same format is also useful to integrate available benchmarks dataset and provide sufficient data to
modern data-driven, large-scale
representation learning methods such as deep learning.

\item \emph{Protocol for representation evaluation}.
There is a strong need of a protocol to benchmark skeleton-based human representations,
which must be independent of learning and application-level evaluations.
Although the representations have been qualitatively assessed
based on their characteristics (e.g., scale-invariance, etc.),
a beneficial future direction is to design quantitative evaluation metrics to
facilitate evaluating and comparing the human representations.

\item \emph{Automated skeleton-based representation learning}.
Deep learning and multi-modal feature learning have recently shown compelling performance in a variety of computer vision and machine learning tasks,
but are not well investigated in skeleton-based representation learning
and can be a promising future research direction.
Moreover, as human skeletal data contains kinematic structures,
an interesting problem is how to integrate this structure as a prior in representation learning.

\item \emph{Real-time, anywhere skeleton estimation of arbitrary poses.}
Skeleton-based human representations heavily rely on the quality of 3D skeleton tracking.
A possible future direction is to extract skeleton information of unconventional human poses (e.g., beyond gaming related poses using a Kinect sensor).
Another future direction  is to
reliably extract skeleton information in an outdoor environment
using depth data acquired from other sensors such as stereo vision and LiDAR.
Although recent works based on deep learning \cite{fan2015combining,toshev2014deeppose,tompson2014joint}
showed promising skeleton tracking results,
real-time processing must be ensured for real-word online applications.

\end{itemize}

\section{Conclusion}\label{sec:Conclusion}

This paper presents a unique and comprehensive survey of the state-of-the-art space-time human representations based 3D skeleton data that is now widely available.
We provide a brief overview of existing 3D skeleton acquisition and construction methods, as well as a detailed categorization of the 3D skeleton-based representations from four key perspectives, including information modality, representation encoding, structure and topological transition, and feature engineering. We also compare the pros and cons of the methods in each perspective.
We observe that multimodal representations that can integrate multiple feature sources usually lead to better accuracy in comparison to methods based on a single individual feature modality.
In addition, learning-based approaches for representation construction, including deep learning, unsupervised feature learning and dictionary learning, have demonstrated promising performance in comparison to traditional hand-crafted feature engineering methods.
Given the significant progress in current skeleton-based representations,
there exist numerous future research opportunities,
such as fusing skeleton data with RGB-D images, cross-training,
and real-time, anywhere skeleton estimation of arbitrary poses.

\bibliographystyle{elsarticle-num}
\bibliography{PerceptSkeleton}	

\end{document}